\documentclass{ws-jaic}
\usepackage[square]{natbib}
\usepackage{graphicx}%
\usepackage{multirow}%
\usepackage{amsmath,amssymb,amsfonts}%
\usepackage{amsthm}%
\usepackage{mathrsfs}%
\usepackage[title]{appendix}%
\usepackage{xcolor}%
\usepackage{textcomp}%
\usepackage{manyfoot}%
\usepackage{booktabs}%
\usepackage{algorithm}%
\usepackage{algorithmicx}%
\usepackage{algpseudocode}%
\usepackage{listings}%
\usepackage{hyperref}
\usepackage{lineno}
\usepackage[all]{xy}
\usepackage{bm}
\usepackage{caption}
\usepackage{comment}
\usepackage{soul}
\usepackage[all]{nowidow}
\usepackage{tikz}
\usetikzlibrary{arrows.meta, positioning, shapes.geometric, fit}

\definecolor{green}{rgb}{0.0, 0.5, 0.0}

\makeatletter
\renewcommand{\doi}[1]{%
  \begingroup
    \edef\tempurl{https://doi.org/#1}%
    \href{\tempurl}{\nolinkurl{doi:#1}}%
  \endgroup
}

\renewcommand{\catchline}[5]{}
\def\@clinebuf{}

\renewcommand{\textcolor}[3][]{#3}

\makeatother
\hypersetup{ colorlinks=true, linkcolor=blue, urlcolor=blue, citecolor=blue }

\begin{document}

\markboth{Robert Prentner}{Categorical AI phenomenology}

\catchline{0}{0}{0000}{}{}

\title{Categorical AI phenomenology: A first-person approach}

\author{Robert Prentner\footnote{ORCID-ID: 0000-0003-1890-0827; please cite the published article: R. Prentner (2026). Categorical AI Phenomenology: A First-Person Approach, \textit{Journal of Artificial Intelligence and Consciousness}, \url{https://doi.org/10.1142/S2705078526500013} }}

\address{Institute of Humanities, ShanghaiTech University,\\
No. 393 Middle Huaxia Road, Shanghai, 201120, China\\
\&\\
Association for Mathematical Consciousness Science,\\
Geschwister-Scholl-Platz 1, Munich, 80539, Germany\\
robert.prentner@amcs.science}

\maketitle


\begin{abstract}
This paper develops a phenomenology‑first approach to artificial consciousness by reframing consciousness as the subjective experience enacted through an agent’s interface with the world. We shift the methodological focus to first‑person structures, modeled mathematically by \textcolor{magenta}{categories derived from} $\bm{Q}$-networks to capture actions and phenomenological invariants. In this framework, $\bm{Q}$-networks are conceptualized as relational interfaces encoding agent-world interaction, analogous to how the dynamical states of a computer depend on its sensory inputs, previous states, and actions. 

Our work provides a rigorous framework for \textit{interface consciousness} to describe computational systems that embed information-processing into phenomenological structure. The approach aligns with 4E approaches to cognition by emphasizing enactive, embedded, and extended dimensions of experience. The paper thus offers a principled, relational, and phenomenological account of artificial phenomenology grounded in categorical mathematics.

\keywords{AI consciousness; (applied) category theory; phenomenological interfaces; $\bm{Q}$-networks; \textcolor{magenta}{structural unification}} 
\end{abstract}

\section{First-person computation}
\label{sec:intro}
What does it mean to say that computers could be conscious? Can a machine subjectively experience its surroundings and inner workings? And what is the best way to study this phenomenon? \citep{amcs23, Gamez08, Dehaene17, Bach19, Kanai19,  Ruffini22, vanRullen21, Butlin23, Blum24} We propose that these questions translate to the question of whether computers can have their \textit{own} subjective perspective on the world. And to study this means doing \textit{phenomenology from the machine's point of view}. 

Consciousness science differs from conventional scientific fields such as the study of chemical reaction networks or genetic regulation. The core difference is that studying consciousness, unlike the latter subjects, requires one to integrate a first-person perspective with conventional third-person science: how is it that it feels like anything for a physical system that processes information, such as a complex molecular system, a human brain, or a computer? This question gave rise to plenty of debates about the underlying metaphysical picture \citep{Chalmers02aa}, which still continue today \citep{Frankish16,Goff19,Friston20,Ellia21,Graziano22,Prentner24c,Kuhn24}. Some of these views seem to make artificial consciousness likely. Others would rule it out. 

In this contribution, by contrast, we want to primarily focus on the \textit{methodological issues} related to the question whether machines could be conscious. Our answer entails a re-conceptualization of consciousness, not as ``inner theater'', but as the \textul{subjective experience enacted through an agent's interface} with the world. Conceivably, such interfaces could be instantiated by various kinds of computational processes, natural and artificial ones. This statement is made more precise in the course of the paper. Talking about agency, at least as it is related to ``enacting a subjective experience,''\footnote{Whether the concept of ``agency'' in fact obviates the possibility of machine consciousness is an important discussion that we do not wish to enter at this point, but see discussions in \citep{Franklin97,Barandiaran09,Roli22,Dung24,Floridi25}.} requires one to consider the first-person perspective. We thus specifically focus on those interfaces that enable an agent to organize computational processes according to phenomenological principles. 

We can first take some inspiration from the scientific study of consciousness in neuroscience \citep{Signorelli21,Seth22,Albantakis25}. It has been proven challenging to exclusively deal with third-person data (such as brain scans) and construct a meaningful science of consciousness. An alternative is to invert the methodology and instead start from the first-person perspective, that is, to begin with an understanding of how appearances are given to us, thus following a “phenomenology-first approach” to consciousness. One model in the neuroscience of consciousness, integrated information theory   \citep{Tononi16,Albantakis23b}, purports to do precisely this:\footnote{Integrated information theory presents a notable methodological advance, but in our opinion the methodology needs to be pursued with even more candor.} 
start with basic, almost self-evident and indubitable, introspective features of conscious experience, and infer how these are in turn represented in certain physical (neural) systems.

We argue that we have to go even deeper, beyond the confines of viewing introspection as if it were just a perception of ``inner objects." We need to let the ``subject speak'' and understand how consciousness shapes the experienced world around us. Arguably, the most systematic philosophical tools for this are provided by phenomenology \citep{Gallagher08,Yoshimi16,sep-phenomenology}, the study of subjective experience as it was conceived by Edmund Husserl and others in the early 20th century. 

%
At the same time, we also want to adopt a mathematical method \citep{Prentner25a}. Eventually, our goal is to integrate first-person studies with contemporary models of computation. \textcolor{magenta}{By doing this, we inevitably reach beyond classical Husserlian phenomenology. Husserl, for example, emphasized that consciousness lacks the exactness of mathematical objects (Ideas I, §73-75), and his analyses were often descriptive rather than formal. Our proposal should therefore be understood as a philosophically-informed extension of these insights, not a reconstruction of Husserl's own position together with a faithful translation into the language of computer science.} 

The specific mathematical framework chosen for this work is \textit{category theory} \citep{Lawvere09,Maclane98}. While highly abstract, category theory has already been applied to theoretical computer science and programming \citep{Milewski19}, but also implications for the study of consciousness have been hinted at (some examples are listed in \citep{Signorelli20a,Tull21,Tsuchiya21,Taguchi23,Prentner24b}).

Any mathematical framework for studying consciousness should start from the fact that we undergo experiences and can relate some of those experiences. Our hypothesis is that subjectivity lives in the ever-changing relations between different experiences. It is the task of (mathematical) phenomenologists to elucidate how this process unfolds in a systematic way.  
Our specific choice of category theory is motivated by the idea that phenomenology is relational, which is illustrated by at least the following: (i) many traditional phenomenological concepts such as “intentionality” or “constitution” are inherently relational in spirit \textcolor{magenta}{\citep{Taguchi19,Voltolini24,Prentner25a}}, often correlating a subjective act and their corresponding objects, and (ii) phenomenological analyses often proceed in terms of relating potential and actual experiences that underlie our manifest (i.e., introspectively available) experience  \citep{Yoshimi16,Prentner25a}. Such an approach to phenomenology can be distinguished from a purely descriptive one that equates the study of consciousness with introspection. 

The key contribution of this paper is to introduce \textul{$\bm{Q}$-networks} as a minimal formalism for modeling first-person structures of experience,\footnote{\textcolor{magenta}{It is important to clarify that “$\bm{Q}$-networks” as used here are not specifically related to Q-learning or Q-function approximators as commonly understood in the reinforcement learning community (see, e.g., \cite{Sutton18}).}} thereby providing a rigorous framework for artificial phenomenology. Our project can be seen as an elaboration on previous work on ``phenomenal spaces'' \citep{Prentner19a} to mathematically treat spaces of experience, enriched by insights from above-mentioned work on category theory in consciousness studies and phenomenology. 

\begin{figure}[hbt]
\centering
\includegraphics[width=0.9\textwidth]{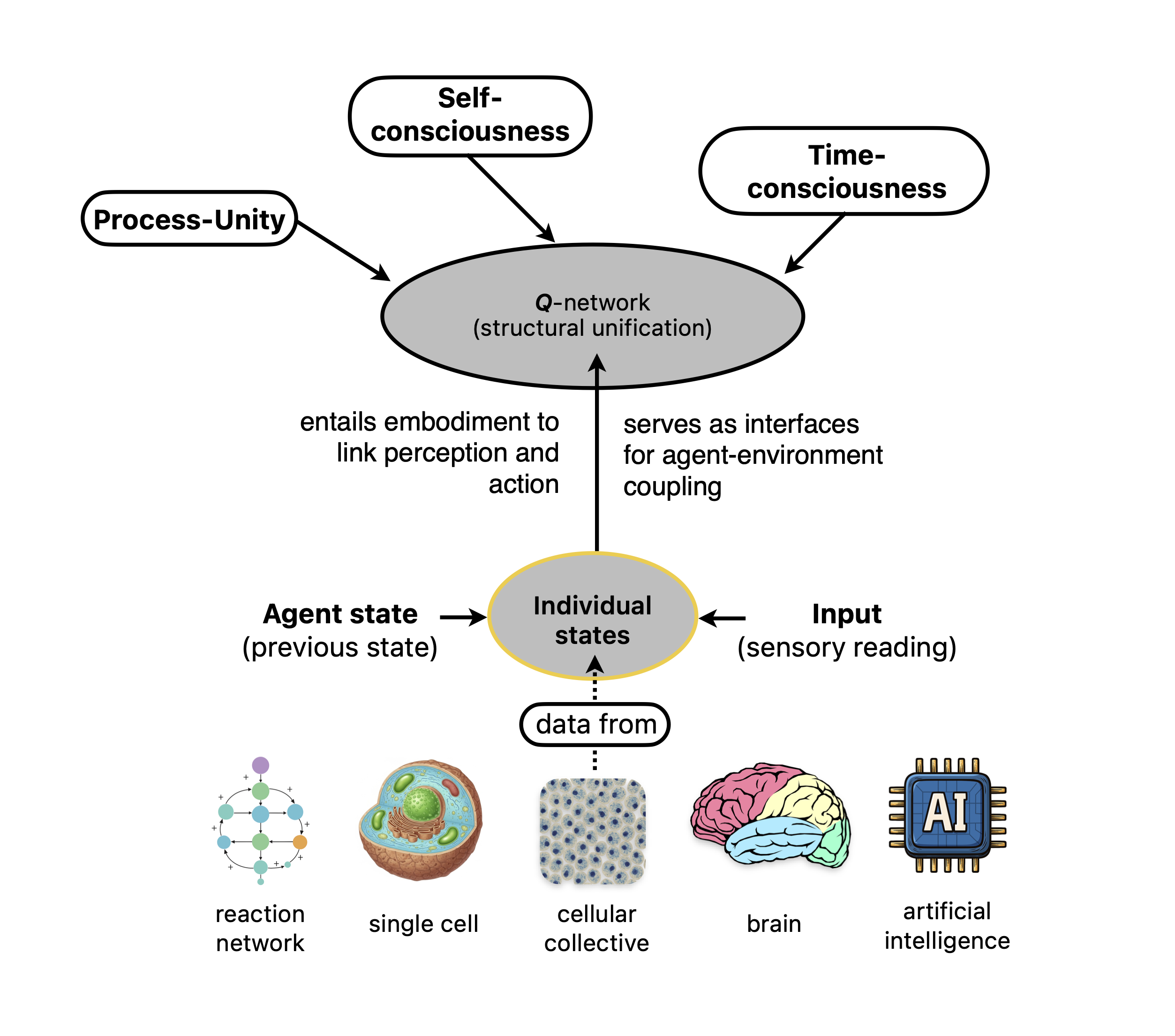} 
\caption{The ``phenomenological hierarchy'' of this paper: data about various natural and artificial information-processing systems (e.g. reaction networks, single cells, cellular collectives, brains or AIs) can be embedded into $\bm{Q}$-networks that encode relations between actual and potential \textcolor{magenta}{states of an agent}. $\bm{Q}$-networks can be seen as exemplars of the larger class of \textit{interfaces} encoding agent-environment interaction, inherently involving embodiment. A phenomenological approach investigates how such interfaces could mediate \textcolor{magenta}{the structural conditions for} subjectivity, potentially giving rise to more complex phenomenological properties mirrored in, for example, process-unity, time-, or self-consciousness. Collectively this defines how \textit{subjective experiences are enacted through an agent's interface with the world}. (Diagram created with the help of Microsoft Copilot.)}
\label{fig:hierarchy}
\end{figure}

 $\bm{Q}$-networks and their relation to category theory are discussed in more detail in in section \ref{sec:phenomenology}, \textcolor{magenta}{where it is proposed to derive categorical structures from them for phenomenological analyses}. We briefly discuss the relation of this approach to 4E cognition and, more generally, to the project of ``interface consciousness'' in section \ref{sec:interface} (including a computational toy model), before we \textcolor{magenta}{revisit phenomenology in section \ref{sec:reflective} to discuss further organizational principles of $\bm{Q}$-network-derived structures. We then} conclude in section \ref{sec:conclusion}.\\ 
The ``phenomenological hierarchy'' laid out in this paper can be inferred from Fig.~\ref{fig:hierarchy}.

\clearpage

\section{Categorical phenomenology}
\label{sec:phenomenology}
\subsection{A doubly relational structure: $\Phi = R^2$}

Before we appeal to the technical specification of our model, we first want to highlight two senses in which phenomenology should be understood as being relational. This further motivates our use of a particular mathematical formalism (i.e., category theory). 

The first sense, in which phenomenology is relational, is hardly controversial to researchers familiar with the historical material since it involves many of the traditional concepts employed in almost any phenomenological investigation, albeit sometimes differing in terminology: 

\begin{itemize}
\item Intentionality,
\item Constitution,
\item Self,
\item Time-consciousness,
\item Intersubjectivity, 
\item[$\vdots$]
\end{itemize}


Another, perhaps less obvious, sense in which phenomenology is primarily relational involves the question of potentiality. Accordingly, phenomenology is an analysis not of actual conscious experiences (though this is undoubtedly their starting point) but of the relation between actual and possible experiences. In the \textit{Crisis}, Husserl wrote that “[each object that I experience] indicates an ideal general set of actual and possible experiential manners of givenness.” \citep{Husserl70} 

\cite{Yoshimi11} provided a physics-metaphor for this. Experiences correspond to points in a state space, and traditional phenomenological analyses are (mostly qualitative) statements about possible trajectories through these spaces. Each trajectory thus connects countless potential experiences and defines a common relational property (technically, a trajectory can be defined as a one-parameter family of points).  

We now further generalize this idea to \textcolor{magenta}{what we call} a minimal mathematical model of phenomenology. Whereas this requirement of minimality follows a general advice for scientific modeling known as Occam's razor, it strikes us as particularly relevant when scientifically studying consciousness, which has a public-facing image of being ``unmathematizable in principle.'' We disagree. Let us list some  basic desiderata of such a mathematization: 
\begin{enumerate}
\item It should start with some (but not too many) explicit assumptions and stay general enough to allow for very different formal statements about experience \citep{Signorelli25}. 
\item It is universal \citep{Kanai24} yet metaphysically neutral \citep{Chen23}. Hence, it could be used to understand the minds of a wide variety of entities, not just humans or animals, but also AIs or aliens. 
\item	It is amenable to the idea that a phenomenological system can reflect on itself, which is further related to the notion of ``self-consciousness'' \citep[cf. also Section \ref{sec:reflective} of the present paper]{sep-self-consciousness-phenomenological}.
\end{enumerate}

We, thereby, should base it on as few essential properties of consciousness as possible. These features are typically modeled on how \textit{our} (limited form of) consciousness appears to \textit{us} in introspection as (limited) observers. While it may be tempting to assume that we have transparent access to our own conscious mind, a substantial body of philosophical and psychological research shows otherwise: humans are notoriously unreliable at reporting what they just perceived, how they achieved their goals, or why they acted or desired in a particular way \citep{Nisbett77,Usher23}. This tendency toward confabulation and misreporting has a striking parallel in large language models, which often generate plausible but inaccurate self-explanations of their reasoning \citep{Stechly25,Shojaee25}. We thus stick to a very limited amount of (highly general) fundamental properties, and in turn reconstruct ``typical'' features of consciousness, such as unity or selfhood.

\textcolor{magenta}{A historical antecedent for our structure‑first approach is Husserl’s early (``pre-transcendental'') work on the formal character of mathematical objects. (See \cite{Husserl03} for the original discussion and \cite{Hartimo10} for recent scholarship on the historical relation between Husserl and mathematics.) For example, Husserl’s notion of a ``definite manifold'' emphasizes the primacy of relations and structures over particular contents: what matters is the formal pattern that can be instantiated across different ontological domains. This resonates with our methodological stance of deriving categorical structures from relational data rather than privileging any particular substrate (whether brain or AI). This further motivates looking for sources of structure that can be abstracted and compared via tools from category theory. 
}

A category is defined by (Def. \ref{def:category}):
\begin{enumerate}
\label{def:category}
\item A collection of objects.
\item For any two objects, a collection of morphisms between them.
\item For each pair of morphisms, there exists a composite morphism if the target object of the first equals the source object of the second, i.e., $f: x \rightarrow y,\ g:y\rightarrow z \Rightarrow g \circ f: x \rightarrow z$. Composition is also associative.
\item For each object, there exist an identity morphism from the object to itself, $1_x$, which satisfies unit laws, i.e., $f \circ 1_x = 1_y \circ f = f.$
\end{enumerate}
In category theory, since there must be a unique \textcolor{magenta}{identity} morphism to any object, we could replace talk about objects by talk about morphisms. Indeed, because of compositionality, we could go even one step further and conceive of the objects as mere interfaces for how morphisms should be composed. 

In our model, we interpret \textcolor{magenta}{phenomenological objects as structures derived from relational data (e.g., cliques, trajectories, or other higher‑order patterns),} irrespective of whether those data come from biological or artificial systems. For example, such an object could \textcolor{magenta}{be derived from} \textcolor{magenta}{a network of possible} sensory states of a cell, a subset of brain states, or some of the memory states in a computer. Morphisms correspond to relations between \textcolor{magenta}{these phenomenological objects.} Our (``intentional'') form of consciousness is then but a special way of relating them.

\subsection{Categories and graphs}
\label{subsec:deriving}
A simple construction \textcolor{magenta}{proceeds from a relational set of states}:%
\footnote{The mathematical model described here is similar to that of \cite{Hoffman23}, who developed a model of consciousness as simplification of a previous theory by \cite{Hoffman14}. The original ``conscious agent theory'' models consciousness as network of conscious agents which are each specified via a 6-tuple, $\langle X,G,\bm{P},\bm{D},\bm{A},n\rangle$: two measurable spaces for the experience and actions ($X$ and $G$) of an agent, three kernels that relate them to themselves and the world, and an integer number that counts kernel executions. Note that we can recover our formalism after choice of $R$ and sufficient integration. The counter $n$ is thereby being neglected.\\
A historical precursor is the \textit{Monadology} of \cite{Leibniz05}.}

\begin{enumerate}

\item At any moment, there are \textcolor{magenta}{potential experiential states,} represented as elements $x \in X$.
\item \textcolor{magenta}{States stand in relations to one another, encoded by a function 
\begin{equation}
\bm{Q}: X \times X \textcolor{magenta}{\rightarrow R}.
\end{equation}
\item The choice of $R$ determines whether the relation is probabilistic, weighted, or binary.} We represent this by a \textit{kernel} $\bm{Q}$\textcolor{magenta}{, which in the probabilistic\footnote{So one can meaningfully talk about \textit{this} \textcolor{magenta}{state}, as opposed to \textit{that} one, $x_i \neq x_j$, at least with a certain probability. Formally, this requires us to postulate that $X$ is measurable. To keep things simple, we assume that $X$ is finite and discrete, and that each $x_j$ just picks out exactly one element of the set. In other words, the relevant algebra of events is the power set $\mathcal{P}(X)$ and the kernel is Markovian (normalized either row‑ or column‑wise, depending on convention).} case assigns transition probabilities, but may also encode weighted or binary relations depending on the application.}
\end{enumerate}

\textcolor{magenta}{ 
}
%
\textcolor{magenta}{We call the resultant object a ``$\bm{Q}$-network.'' The definitional flexibility is intentional. Phenomenology requires only that experiential states stand in relations; it does not commit us to a specific numerical type. Temporal scaffolding, similarity structure, or entailment are all admissible relational forms. What makes a structure a ``$\bm{Q}$‑network” is that the kernel $\bm{Q}$ encodes how \emph{potential experiential states are related}, from which phenomenological objects can later be derived.}

We thereby specifically focus on (i) \textit{relations of states}, and (ii) the influence of \textit{action} \textcolor{magenta}{on those, for example, by discussing action-indexed families of such kernels.} 

We illustrate this with the simple case where relations between \textcolor{magenta}{states} can be described deterministically and where the total number of \textcolor{magenta}{states} is finite. In this case, any kernel can be written as a finite \textcolor{magenta}{(binary)} matrix with entries zero or one. This need not be the case more generally.

We interpret relations as the possibility of one \textcolor{magenta}{potential experiential state} being followed by another: $x_j$ relates to $x_k$ if and only if $x_j$ is followed by $x_k$ according to the kernel $\bm{Q}$, for example, modeling an attentional process or sensory dynamics.%
\footnote{Note that this suggests an interpretation on the level of “earlier/later,” which presupposes an ordering scheme that might not yet be given \textit{from the perspective of the system} itself but only in the external observer's spatiotemporal framework. However, we will adopt this interpretation for what follows for simplicity and concreteness. The reader is advised to keep in mind that the kernel more generally expresses a (potentially asymmetric) relation between different (actual or potential) experiences. Other notable interpretations are similarity-structures (symmetric) or entailment relations (asymmetric). For a general discussion on the varieties of the notion of ``structure'' as used in consciousness studies, see \cite{Kleiner24}.}  

It is instructive to first look at a system that has only two possible states, \textcolor{magenta}{$x_0$ and $x_1$. A kernel $\bm{Q}$ on this set is a $2\times 2$  matrix that has the form 
\begin{equation}
\bm{Q}=\begin{bmatrix} \bm{Q}_{00} & \bm{Q}_{01} \\ \bm{Q}_{10} & \bm{Q}_{11} \end{bmatrix}, 
\end{equation}
which corresponds to the following graphical representation:}
\begin{equation}
\nonumber
\xymatrix{
*++[o][F-]{x_0} \ar@/^1pc/[rr]^{\bm{Q}_{10}} \ar@(ul,dl)_{\bm{Q}_{00}}  && *++[o][F-]{x_1} \ar@/^1pc/[ll]^{\bm{Q}_{01}} \ar@(dr,ur)_{\bm{Q}_{11}} 
}
\end{equation}
 A possible choice of $\bm{Q}$ would be realized by the ``$\bm{not}$-kernel'', 
$\bm{Q} = \begin{bmatrix} 
0& 1 \\1 & 0 
\end{bmatrix}$, where $x_0$ is followed by $x_1$, and vice versa. Another choice would be the kernel that finds that any \textcolor{magenta}{state} will lead to the same \textcolor{magenta}{state}, say $x_0$, regardless of where it started, $\bm{Q} = \begin{bmatrix} 
1& 1 \\0 & 0 
\end{bmatrix}.$ 
\textcolor{magenta}{Under the constraints that (i) each row contains exactly one ‘1’ (deterministic transition) and (ii) every state has at least one outgoing transition,} there are four possible kernels, corresponding to the matrices $\left\{ \bm{id},\bm{0},\bm{not},\bm{1} \right\} = \left\{ \begin{bmatrix} 1& 0 \\0 & 1 \end{bmatrix}, \begin{bmatrix} 1& 1 \\0 & 0 \end{bmatrix}, \begin{bmatrix} 0& 1 \\1 & 0 \end{bmatrix},  \begin{bmatrix} 0& 0 \\1 & 1 \end{bmatrix}  \right\}.$ 

As an illustration, one could think of a robot with a single memory state. The transition between an ``off'' and an ``on'' state could be represented by the matrix $\begin{bmatrix} 0&1 \\ 1&0 \end{bmatrix}$. Whenever the state is off now, it will be on in the future; and vice versa (discounting for any external effects.) 

Is it fair to say that such a system would be ``minimally conscious'' as in the example of the photodiode in \citep{Oizumi14}? The system lacks \textit{practically any} interesting phenomenological structure, so this question is hard to answer. More generally, we would maintain that it is about uncovering \textit{relations} between networks, not the intrinsic properties of networks that let us connect this to phenomenology.   
The question of AI consciousness is more akin to the question \textit{to what degree} does the computer exhibit certain relational structures, rather than the question whether it is endowed with a ``magic sauce'' (that we anyway cannot hope to define).

Concretely, one may study kernels on a set of $n$ elements, for which there are $n^n$ (deterministic) kernels. Among these, exactly $n!$ are invertible, corresponding to the permutations of the set $\{1,2,\hdots,n\}$. Within this class, one can further classify kernels according to cycle components, i.e. the disjoint cycles in the permutation decomposition.

For example, a $\bm{Q}$-network with $n=3$ elements may admit an invertible kernel that decomposes into two components (one loop and one singleton). A representative case is shown below:
\begin{equation}
\nonumber
\xymatrixcolsep{1.25cm}
\xymatrixrowsep{.5cm}
\xymatrix{
*++[o][F-]{x_0} \ar@/^1pc/[dd]^{\bm{Q}_{10}=1} \\
& *++[o][F-]{x_2} \ar@(dr,ur)_{\bm{Q}_{22}=1} && \bm{Q} = \bm{not} \otimes \bm{1}\\
 *++[o][F-]{x_1} \ar@/^1pc/[uu]^{\bm{Q}_{01}=1} 
}
\end{equation}
Table \ref{tab:clusters} summarizes the distribution of invertible kernels by cycle components for small $n$. Results for larger $n$ can be obtained combinatorially, and efficient algorithms exist for computing the cycle decomposition of arbitrary kernels.

\begin{table}
\centering
\captionsetup{width=.8\textwidth}
\caption{Number of invertible kernels on $n$ elements with at least one cycle of size $k$. The notation $(c_j,c_k,\cdots)$ indicates the sizes of cycles.}
\begin{tabular}{lll}
$nk$ & $(c_j,c_k,\cdots)$ & number of kernels\\
\hline
\multicolumn{3}{l}{$n =3$} \\
\hline
31 & (1,1,1) & 1 \\
32 & (2,1) & 3 \\
33 & (3) & 2 \\
total: && $3! = 6$ \\
\hline
\multicolumn{3}{l}{$n =4$} \\
\hline
41 & (1,1,1,1) & $1$ \\
42 & (2,2), (2,1,1) &  ${4 \choose 2}/2 + {4 \choose 2} = 9$ \\
43 & (3,1) & $2 \cdot {4 \choose 3} = 8$ \\
44 & (4) & $3! = 6$\\
total: && $4! = 24$\\
\hline
\multicolumn{3}{l}{$n =5$} \\
\hline
51 & (1,1,1,1,1) & $1$ \\
52 & (2,2,1), (2,1,1,1) & $3 \cdot {5 \choose 1} + {5 \choose 3} = 25$ \\
53 & (3,2),(3,1,1) & $2 \cdot {5 \choose 2} + 2 \cdot {5 \choose 2} = 40$\\
54 & (4,1) & $6 \cdot {5 \choose 1} = 30$ \\
55 & (5) & $4! = 24$  \\
total: && $5! = 120$\\
\hline
\vdots
\end{tabular}
\label{tab:clusters}
\end{table}  

Another feature of this representation, which also aligns with the potentiality-interpretation of phenomenology, is that actions (such as movements) can be taken to \textit{parametrize} the space of possible relations between \textcolor{magenta}{potential experiential states}. We assume that the specific subjective perspective encoded by the $\bm{Q}$-network depends not only on what states are currently available to the system but also, crucially, on the way it could transition between them upon acting a certain way. One could, for example, represent our simple system as a ``monoid'', illustrating a collection of four different kernels from a set to itself:
\begin{equation}
\nonumber
\xymatrix{
*++[o][F-]{X} \ar@(ul,dl)_{\bm{id}} \ar@[magenta]@(dr,ur)_{\bm{not}} \ar@[cyan]@(ur,ul)_{\bm{1}} \ar@[orange]@(dl,dr)_{\bm{0}}
}
\end{equation}
or, written a little less concisely: 
\begin{equation}
\nonumber
\xymatrix{
*++[o][F-]{x_0} \ar@[magenta]@/^2pc/[rr]^{\bm{not}_{0}} \ar@[cyan]@/^/[rr]^{\bm{1}_{0}} \ar@(ul,dl)_{\bm{id}_{0}} \ar@[orange]@(dl,dr)_{\bm{0}_{0}}  && *++[o][F-]{x_1} \ar@[magenta]@/^2pc/[ll]^{\bm{not}_{1}} \ar@[orange]@/^/[ll]^{\bm{0}_{1}} \ar@(dr,ur)_{\bm{id}_{1}} \ar@[cyan]@(ur,ul)_{\bm{1}_{1}}
} 
\end{equation}
Colors indicate one of the four possible kernels. Categorical phenomenology is typically based on the co-presence of many such kernels. Indeed, as the number of \textcolor{magenta}{states} grows, the number of possible kernels quickly becomes very large. 

\textcolor{magenta}{In order to move the discussion from the analyses of state transitions to phenomenology, we appeal to the} framework of category theory \citep{Lawvere09,Maclane98,Ehresmann07}. \textcolor{magenta}{Note that we do not treat the $\bm{Q}$-network itself as a category (e.g., via the free-category construction). Instead, we investigate categories derived from $\bm{Q}$-networks, depending on the structural features under consideration.%
\footnote{\textcolor{magenta}{A brief phenomenological-methodological remark: the transition from $\bm{Q}$-networks to derived categorical structures can be read in both a static and a genetic sense. In the static reading, a $\bm{Q}$-network specifies structural relations among states that let us derive categorical objects, interpreted to be related to intentional experience. In a more genetic reading, such relations can be interpreted as giving rise to higher-order experiential structures, echoing the distinction in phenomenology between descriptive (static) analyses and those that trace how higher-order acts emerge from lower-order ones. An explicitly genetic interpretation is an interesting direction for future work.}}
In the toy model of Section \ref{subsec:toy-models}, for example, objects are cliques and morphisms are inclusions, yielding a poset-like category. More generally, one should distinguish between the transition graph of states and the categorical structures derived from it.} 

Category theory provides tools that go beyond standard graph theory, allowing us to capture, for example, invariants, actions, and various kinds of universal phenomenological structures:

\begin{enumerate}
\item \textit{Invariants}. Phenomenologists often look for structures that remain stable across different modes of experience. One way, already emphasized by \cite{Husserl12}, proceeds by systematically uncovering such invariants (equivalence classes). 


\textcolor{magenta}{For example,} invariants can then be identified by functorial correspondences that preserve structure across thresholds. In a practical example (compare Section \ref{subsec:toy-models}), this can amount to computing persistent homology of $\bm{Q}$‑network-derived structures. \textcolor{magenta}{As detailed in this section, such invariants provide summaries of the global structure of phenomenological objects, for instance, via Betti numbers that track connectedness and cyclic organization.}

\item \textit{Actions}. Beyond perceptual invariants, phenomenology also studies how consciousness changes under action. One example from phenomenology is Husserl’s method of variation in imagination, where experience changes upon mentally changing some of its properties, e.g., inverting the colors of a picture we see before our mind’s eye. Some variations might also happen if we physically engage in certain activities. Related, in models of “active inference”  \citep{Clark17,Parr22}, it is assumed that what an organism perceives depends on the way it acts in the world. In applications to computational systems, one might look for the invariants that the computer ``sees'' as a result of a computational process. 

One could model such action with functors that act on categories. A functor can be described as a mapping from objects in one category to objects in a target category (including the category itself):
\begin{equation}
F: \mathcal{C} \rightarrow \mathcal{D},
\end{equation}
Thereby, functorial mappings preserve categorical structure, such as composition and identity laws. In particular, it is the case that if a morphism exists in $\mathcal{C}$, it also exists in $\mathcal{D}$:
\begin{equation}
A \rightarrow B \Rightarrow F(A) \rightarrow F(B).
\end{equation}

Moreover, actions can be composed: sequential functors correspond to sequential transformations of experiential structure. Fig. \ref{fig:action} illustrates this with abstract nodes $A,B,C$, emphasizing that we are now working at the categorical level (functors between categories), rather than at the level of the individual states $x_i$ of a $\bm{Q}$‑network. A particularly simple type of action is represented by functors corresponding to ``loosening'' and ``sharpening'' attention (cf.  the toy model in Section \ref{sec:interface}).

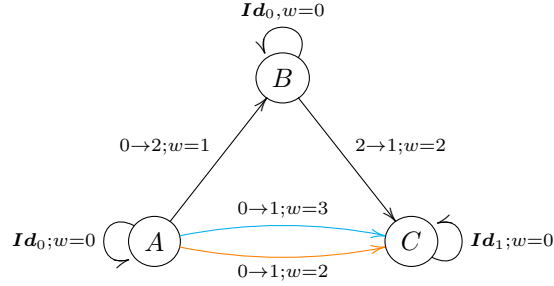
\begin{figure}[htb]
\begin{equation}
\nonumber
\xymatrixcolsep{1.cm}
\xymatrixrowsep{1.5cm}
\xymatrix{
& *++[o][F-]{B} \ar@(ur,ul)_{\bm{Id}_{0},w=0} \ar[dr]^{2 \rightarrow 1; w=2}  & \\
*++[o][F-]{A} \ar[ur]^{0 \rightarrow 2; w=1}  \ar@[cyan]@/^/[rr]^{0 \rightarrow 1; w=3} \ar@[orange]@/_/[rr]_{0 \rightarrow 1; w=2} \ar@(ul,dl)_{\bm{Id}_{0}; w=0}  && *++[o][F-]{C}  \ar@(dr,ur)_{\bm{Id}_{1}; w=0} 
} 
\end{equation}
\caption{From $A$ to $B$ via $C$, one can compose the mapping in two distinct ways. In one case, the weight of the composite is the sum of the parts (cyan); in another, it is the maximum (orange). This flexibility illustrates how categories can accommodate different rules for composing actions. (This example has been inspired by the example of a transport-network discussed in \citep{Ehresmann07}).}
\label{fig:action}
\end{figure}

\item \textcolor{magenta}{\textit{Structural conditions} for phenomenological unification}. Category theory often appeals to universal constructions, for example (co)limits \citep{nlab:colimit}. Concretely, a colimit unifies a pattern of relations into a single minimal object through which all such relations factor. Fig. \ref{fig:colimit} illustrates this idea: the colimit $C$ serves as the minimal cone through which all morphisms from the pattern must pass. In the colimit diagram, the objects $X$, $Y$, $W$, $Z$ represent local sub‑structures \textcolor{magenta}{of the category}, together with the mappings between them; \textcolor{magenta}{the labels $w$ denote arbitrary weights assigned to arrows to demonstrate different composition rules (e.g., additive vs. max‑composition).} The colimit $C$ then serves as the minimal unifying object through which all such relations factor. In the toy model presented in Section \ref{subsec:toy-models}, this abstract idea acquires a concrete computational marker \textcolor{magenta}{when certain Betti numbers collapse or stabilize (e.g., $\beta_0$ reducing to $1$), indicating that disparate relational components have merged into a unified structure.} 
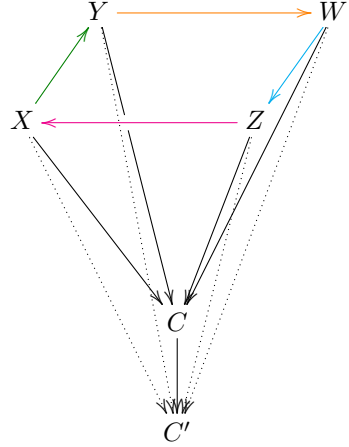
\begin{figure}[htb]
\begin{equation}
\nonumber
\xymatrixcolsep{0.5cm}
\xymatrixrowsep{1.cm}
\xymatrix{
& Y \ar@[orange][rrr] \ar[dddr]|!{[d];[dr]}\hole \ar@{.>}[ddddr] &&& W\ar[dddll]\ar@{.>}[ddddll] \ar@[cyan][dl] \\
X\ar[ddrr]\ar@{.>}[dddrr]\ar@[green][ur] && &  Z\ar[ddl]\ar@{.>}[dddl]\ar@[magenta][lll]  \\\\
&  &C\ar[d] & &\\
&& C' && 
}
\end{equation}
\caption{Illustration of the concept of colimit in a category. All mappings between objects in the pattern (colors) commute with mappings to $C$. Specifically, the colimit, if it exists, picks out the minimal cone $C$. Any mapping from the pattern to another cone $C'$ factors uniquely through $C$ via a mapping from $C$ to $C'$.}
\label{fig:colimit}
\end{figure}
 
In philosophical terms, we can think of the colimit as the unified locus of perception–action loops. Each object in the diagram represents a partial trajectory of experience, parametrized by action, but the colimit gathers them into a single point of convergence. It is the minimal object through which all such loops must pass, and thus encodes the conditions under which perception and action are integrated. In this sense, the colimit functions as the ``meeting point'' of all experiential dynamics, rather than merely summarizing relations. 
Furthermore, this object could itself be accessible to reflection, even if normally not part of immediate awareness, suggesting representation in terms of higher category theory \citep{nlab:higher_category_theory}.
\end{enumerate}
Together, these examples suggest that certain mental properties can be represented categorically and related functorially to physical data, once \textcolor{magenta}{they have been} embedded into $\bm{Q}$-networks. This conception aligns with earlier proposals in the literature \citep{Ehresmann15,Tsuchiya16,Atmanspacher22,Prentner24b}.




\section{Interface consciousness and a concrete toy model}
\label{sec:interface}
\subsection{Interface consciousness}

The outlined approach dovetails with a broader research program on \textit{interface consciousness}, which conceives of consciousness as the enactment of subjective experience through interfaces. $\bm{Q}$-networks provide the formal scaffolding for such interfaces, which are defined by relations between \textcolor{magenta}{the states of an agent}. 

Recall the ``phenomenological hierarchy'' presented in Fig. \ref{fig:hierarchy}. Here we distinguished (i) between \textcolor{magenta}{states}, (ii) relations between \textcolor{magenta}{states} that define $\bm{Q}$-networks, and (iii) networks that are organized according to further phenomenological constraints, which all
\textcolor{magenta}{correspond to different aspects of interfaces. This models how third-person data are embedded into a ``phenomenal space'' \citep{Prentner19a}, how actual and possible states are related to define experience, and how such relations are organized into forms of intentional consciousness.} 

What matters for consciousness is relational, and relations \textit{prima facie} do not dependent on the nature of the substrate they are relating.\footnote{Of course, this leaves open that certain types of relations could only exist between certain types of objects.} By modeling \textit{interfaces rather than substrates}, we align our study with received approaches in the vicinity of 4E cognition \citep{Menary10,Newen18}, applied to consciousness in a non-reductive way. Below are just three examples from the contemporary literature:

\begin{enumerate}  
\item \textit{Active Inference} approaches to consciousness \citep{Seth22,Albarracin22,Ramstead23} find that the content of consciousness is shaped more by an agent's internal beliefs and possible actions than by the ``raw'' sensory signals it receives. From this perspective, one could say that consciousness research is about uncovering the \textit{phenomenological} principles of (Bayesian) active inference \citep{Parr22}. \textit{Subjective experience is about harboring a generative model \textcolor{magenta}{(related to phenomenology)}.}
\item The related \textit{Interface Theory of Perception} \citep{Hoffman15a,Prakash20a,Prakash21,Prentner21,Prentner24c} argues that what an agent perceives is  shaped by utility, not by true environmental structure. The structures we perceive are embedded in an agent-specific way of representing its environment. Applied to consciousness science, this means to understand how those agent-specific ways of representing align with phenomenological ideas. \textit{Consciousness is fundamentally relational.}

\item The \textit{Projective Consciousness Model} \citep{Rudrauf17,Williford22} is premised on the idea that the contents of consciousness (all contents, not only the visual ones) amount to representations that are organized in a projective space. Moreover, the way how those contents are represented accounts for certain phenomenological insights that enables a representational system to carry out specific functions aimed at behavioral control and optimization. \textit{Phenomenology is tied to function.}
\end{enumerate}

\textcolor{magenta}{This relational and projective character of experience has long been emphasized in phenomenology, for example in Aron Gurwitsch's field theory of consciousness \citep{Gurwitsch64} and Alfred Schutz's analyses of anticipatory, action‑oriented structures of experience \citep{Schutz67}.} Therefore, \textit{interface consciousness} is not merely a metaphor but a \textcolor{magenta}{philosophically-informed} research program that can be computationally instantiated and experimentally probed. The following subsection illustrates this with a concrete toy model.

\subsection{A toy interface}
\label{subsec:toy-models}
To illustrate how the abstract framework can be implemented, we sketch a simple computational toy model. The aim is not to capture the full richness of phenomenology, but to demonstrate how third-person data can be embedded into a phenomenal space, structured into $\bm{Q}$-networks, and then analyzed with categorical and topological tools. \textcolor{magenta}{Unlike in the examples of Section \ref{subsec:deriving}, we do not adhere to the requirement that transitions between states are (families of) deterministic kernels (although they could be, such as in the case of temporal adjacency discussed below).}

\subsubsection{Third-person data and phenomenal embedding}

\begin{figure}[hbt]
	\centering
    \includegraphics[width=1\textwidth]{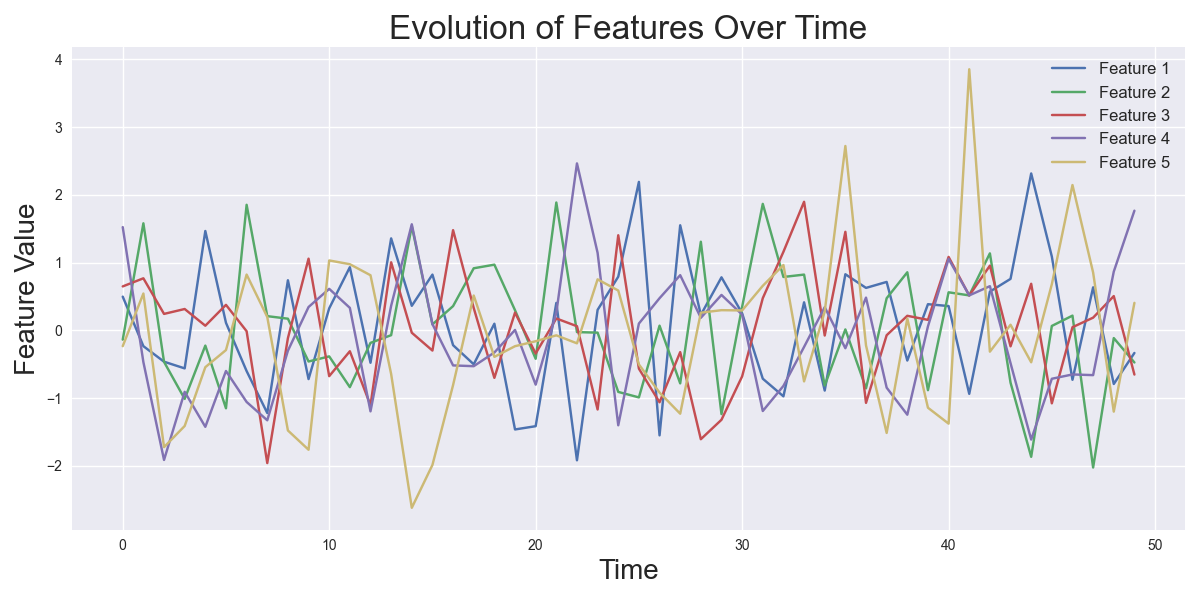}
	\caption{Evolution of features over time (``third-person data'').}
	\label{fig:timeseries}
\end{figure}

We begin with a synthetically generated time series of $n=50$ points, each with $5$ features (Fig.~\ref{fig:timeseries}). \textcolor{magenta}{In a realistic scenario, such features can refer to a wide range of empirical or simulated systems, for example molecular concentrations, sensor readings, neuronal activity patterns, internal memory states of an artificial agent, or higher-level structures such as IIT’s cause–effect repertoires \citep{Tononi16,Albantakis23b}. The specific choice of features is not essential for the present illustration; what matters is that they encode third-person data from which relational structure can be derived.}

\begin{figure}[htb]
    \hspace*{-2cm}
    \includegraphics[width=1.3\textwidth]{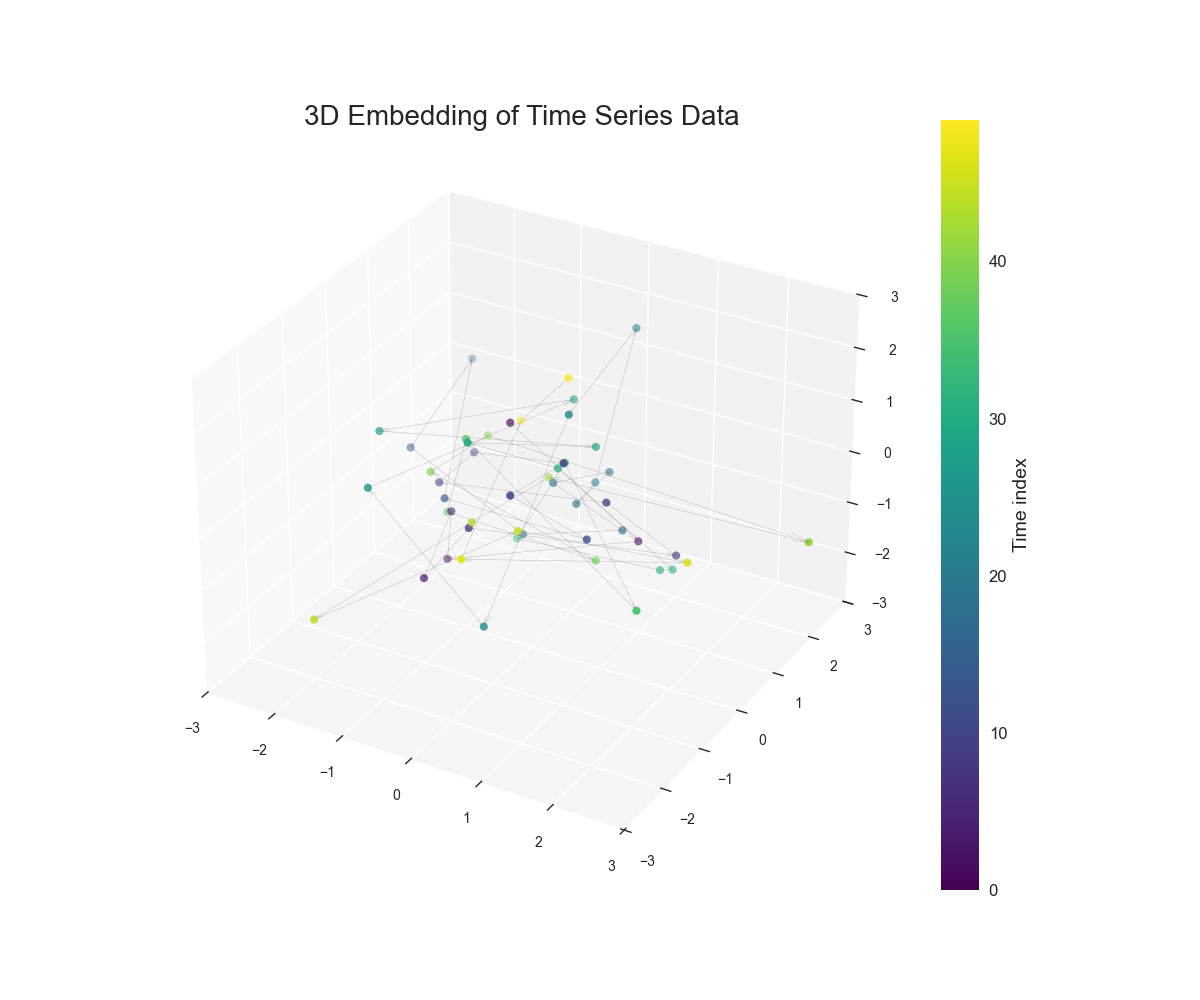}
    \caption{3D embedding of the time series into a phenomenal space. Each point
    corresponds to \textcolor{magenta}{a state of a $\bm{Q}$-network; arrows indicate temporal succession.}}
    \label{fig:embedding}
\end{figure}

These data are then embedded into a three-dimensional space using principal component 
analysis (PCA), yielding a geometric representation in which each point corresponds to a 
potential or actual \textcolor{magenta}{state of the resulting $\bm{Q}$-network (Fig.~\ref{fig:embedding}). More sophisticated versions of the model could employ nonlinear or learned embeddings, such as autoencoders or contrastive neural embeddings.}


\subsubsection{Defining Q-networks}

\begin{figure}[hbt]
    \centering
    \includegraphics[width=1\textwidth]{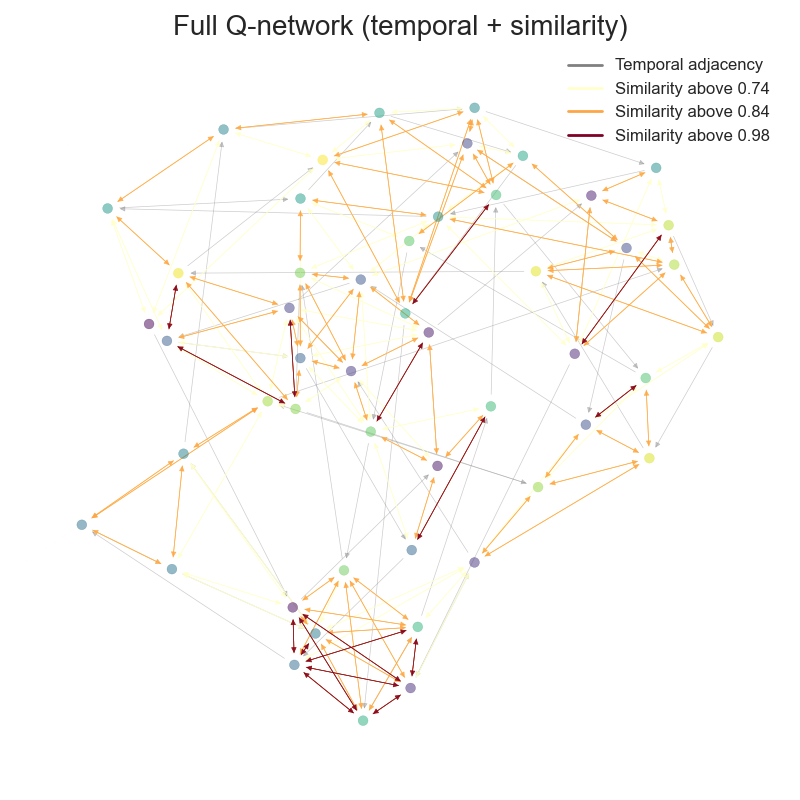}
    \caption{Full $\bm{Q}$-network combining temporal adjacency (gray edges) and
    similarity-based edges (colored). Node color encodes time index.}
    \label{fig:Q-network}
\end{figure}

From the time series embedding, we construct a $\bm{Q}$-network (Fig. \ref{fig:Q-network}) as follows:

\begin{enumerate}
\item Compute pairwise similarity among embedded states $\varphi(s)$, based on the cosine similarity matrix with entries
\begin{equation}
M_{ij}^{\mathrm{sim}} = \frac{\varphi(s_i)\varphi(s_j)}{||\varphi(s_i)||||\varphi(s_j)||}.
\end{equation}
A similarity graph $G_{\textrm{sim}}(\tau)$ is then defined as undirected graph with vertex set $S$ and an (undirected) edge $\lbrace i,j \rbrace$ iff $M_{ij}^{\mathrm{sim}} \ge \tau$, where $\tau \in [-1,1]$ is an arbitrarily chosen similarity threshold. Similarity graphs are later used to compute persistent homology. 
\item Record temporal adjacency between successive states in the time series. A directed temporal adjacency graph is defined if there is an observed transition from state $i$ to state $j$ in the recorded trajectory. 
\end{enumerate}
\textcolor{magenta}{This separation preserves interpretability and allows comparison between hypothesized ``topological phenomenal properties'' (similiarity-based) and observed dynamics (temporal adjacency-based) in future, more realistic settings.}

While recorded separately, both constructions are combined into a single graph for illustration (Fig. \ref{fig:Q-network}), where the similarity-based relation is treated as a bi-directional arrow (weighted by the threshold, indicated by colors).

\begin{figure}[htb]
    \centering
    \includegraphics[width=1\textwidth]{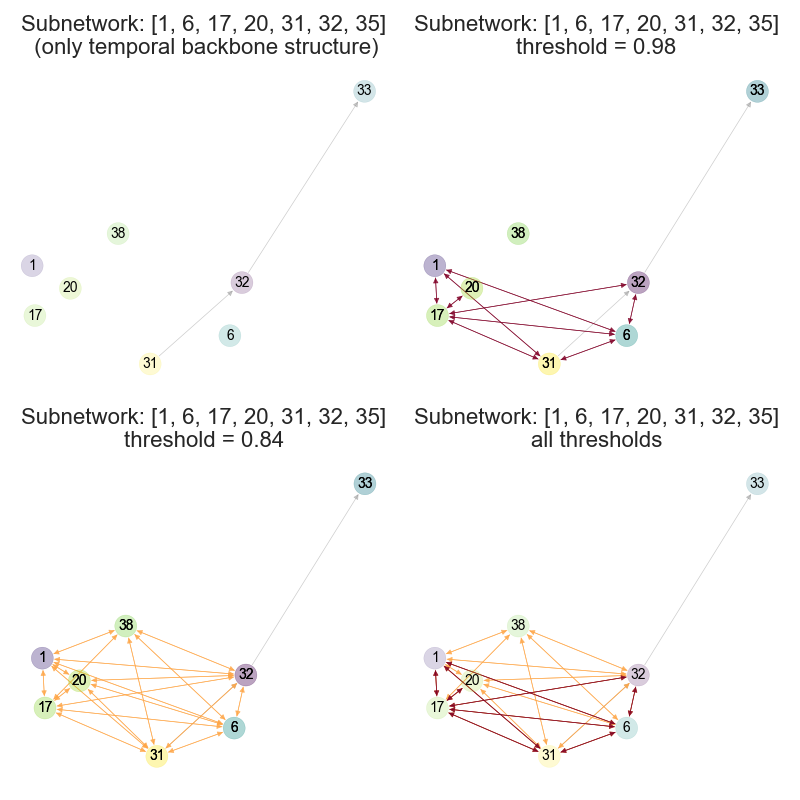}
    \caption{Example subnetworks for selected nodes at different similarity thresholds.}
    \label{fig:subnetwork}
\end{figure}

Subgraphs can be extracted for selected nodes $[1,6,17,20,31,32,35]$ and similarity thresholds (Fig. \ref{fig:subnetwork}), revealing different ``graph motifs'' \citep{Newmans10}. A \textcolor{magenta}{exemplary} discussion of some motifs is provided in Table \ref{tab:motifs}.

\begin{table}[htb]
\label{tab:motifs}
\centering
\caption{Motif analysis of similarity-based $\bm{Q}$-network subgraphs across thresholds.}
\label{tab:motifs}
\begin{tabular}{lccc}
\toprule
\textbf{Threshold} & \textbf{\# Max cliques} & \textbf{Avg clique size} & \textbf{\# Triangles} \\
\midrule
0.98 & 11 & 2.27 & 1 \\
0.84 & 32 & 3.00 & 16 \\
0.74 & 32 & 3.97 & 9 \\
\bottomrule
\end{tabular}

\vspace{0.6em}
\small
\textbf{Examples (representative subsets):}
\begin{itemize}
  \item \textbf{0.98:} Max cliques: [3, 37], [5, 44], [9, 40], [12, 35], [14, 40], [14, 48], [16, 33], [17, 20], [17, 6, 32], [17, 6, 1, 31], [30, 34]. Triangles: [17, 6, 32].
  \item \textbf{0.84:} Max cliques (subset): [1, 32, 6, 17, 31, 20], [1, 32, 6, 17, 31, 38], [2, 16, 33], [2, 3, 37], [3, 8, 13], [3, 8, 37], [4, 24], [4, 44, 5], [5, 43, 44], [7, 41, 46], [8, 15, 9], [8, 15, 13, 39], [9, 40, 48, 14], [9, 40, 15], [10, 34, 30, 23], [10, 11], [11, 45, 47], [11, 29], [12, 46, 41], [12, 46, 35], [18, 21, 42], [18, 21, 19], [22, 26], [22, 28], [23, 49, 34, 30], [25, 43], [25, 27], [26, 48], [27, 39], [28, 34, 30], [29, 36], [43, 45, 47]. Triangles (subset): [2, 16, 33], [2, 3, 37], [3, 8, 13], [3, 8, 37], [4, 44, 5], [5, 43, 44], [7, 41, 46], [8, 15, 9], [9, 40, 15], [11, 45, 47], [12, 46, 41], [12, 46, 35], [18, 21, 42], [18, 21, 19], [28, 34, 30], [43, 45, 47].
  \item \textbf{0.74:} Max cliques (subset): [0, 22, 49], [0, 22, 26], [1, 20, 32, 38, 17, 6, 31], [1, 20, 19, 21], [2, 37, 8, 3], [2, 37, 33, 16], [4, 24, 44, 5], [4, 13], [5, 44, 43, 25], [5, 44, 43, 47], [6, 32, 7, 17, 20], [7, 41, 12, 46], [10, 11, 29], [10, 23, 49, 34, 30], [10, 23, 29], [11, 43, 45, 47], [12, 35, 41, 46], [12, 35, 45], [15, 8, 40, 9], [15, 8, 3, 9, 37], [15, 8, 3, 13, 37], [15, 8, 3, 13, 39], [15, 27, 13, 39], [15, 14, 40, 9], [18, 42, 19, 21], [22, 28, 49], [23, 28, 49, 34, 30], [25, 27, 36], [25, 27, 13, 39], [25, 29, 36], [26, 48, 14], [42, 40, 48, 9, 14]. Triangles: [0, 22, 49], [0, 22, 26], [10, 11, 29], [10, 23, 29], [12, 35, 45], [22, 28, 49], [25, 27, 36], [25, 29, 36], [26, 48, 14].
\end{itemize}
\end{table}
\clearpage

\subsubsection{Categorification and invariants}

In a next step, \textcolor{magenta}{an ``inclusion-category'' should be derived from} the $\bm{Q}$-network. From now on, we focus only on similarity relations and ignore temporal scaffolding \textcolor{magenta}{by only using the undirected similarity graph $G_{\textrm{sim}}(\tau)$. We thereby introduce the notion of a ``clique'' from algebraic topology: a clique is a set of vertices that are pairwise connected in the underlying similarity graph. We use the clique (flag) complex construction, in which cliques of the similarity graph are treated as simplices. This construction is standard in algebraic topology and topological data analysis  (see, e.g., \cite{Edelsbrunner10}). 
To relate this to category theory (chosen to establish the link to phenomenology)},%
\footnote{\textcolor{magenta}{The observations in this subsection are heuristics based on the toy model rather than general mathematical theorems. In particular, the apparent coincidence of Betti-number crossings with qualitative regime shifts in the derived categories is an empirical pattern in this example, not a categorical statement about colimits, universal constructions, or functorial properties. Category theory is introduced here not to establish such correspondences, but because it provides the structural language needed to model how experiential structures relate, transform, and compose. Persistent homology captures the shape of relational data, whereas category theory captures the transformations between such shapes (e.g., inclusions, functors, and action-induced mappings).}}
we treat cliques as objects and inclusions between cliques as morphisms. Parameterized by the similarity threshold, this yields a filtered category \citep{nlab:filtered_category}. Minimal inclusions can be visualized as Hasse diagrams.

\begin{figure}[hbt]
    \centering
    \begin{tabular}{cc}
        \includegraphics[width=0.45\textwidth]{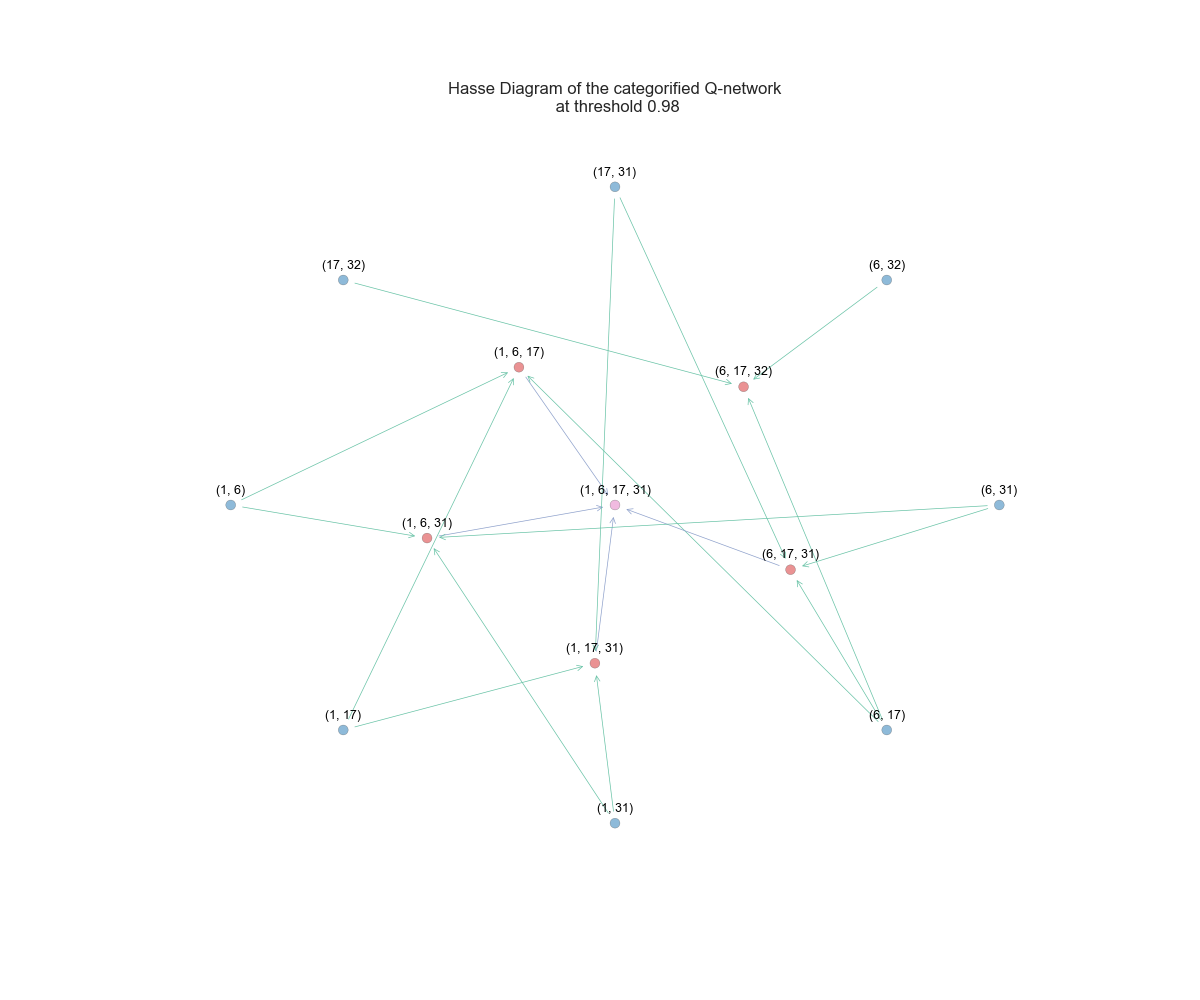} &
        \includegraphics[width=0.45\textwidth]{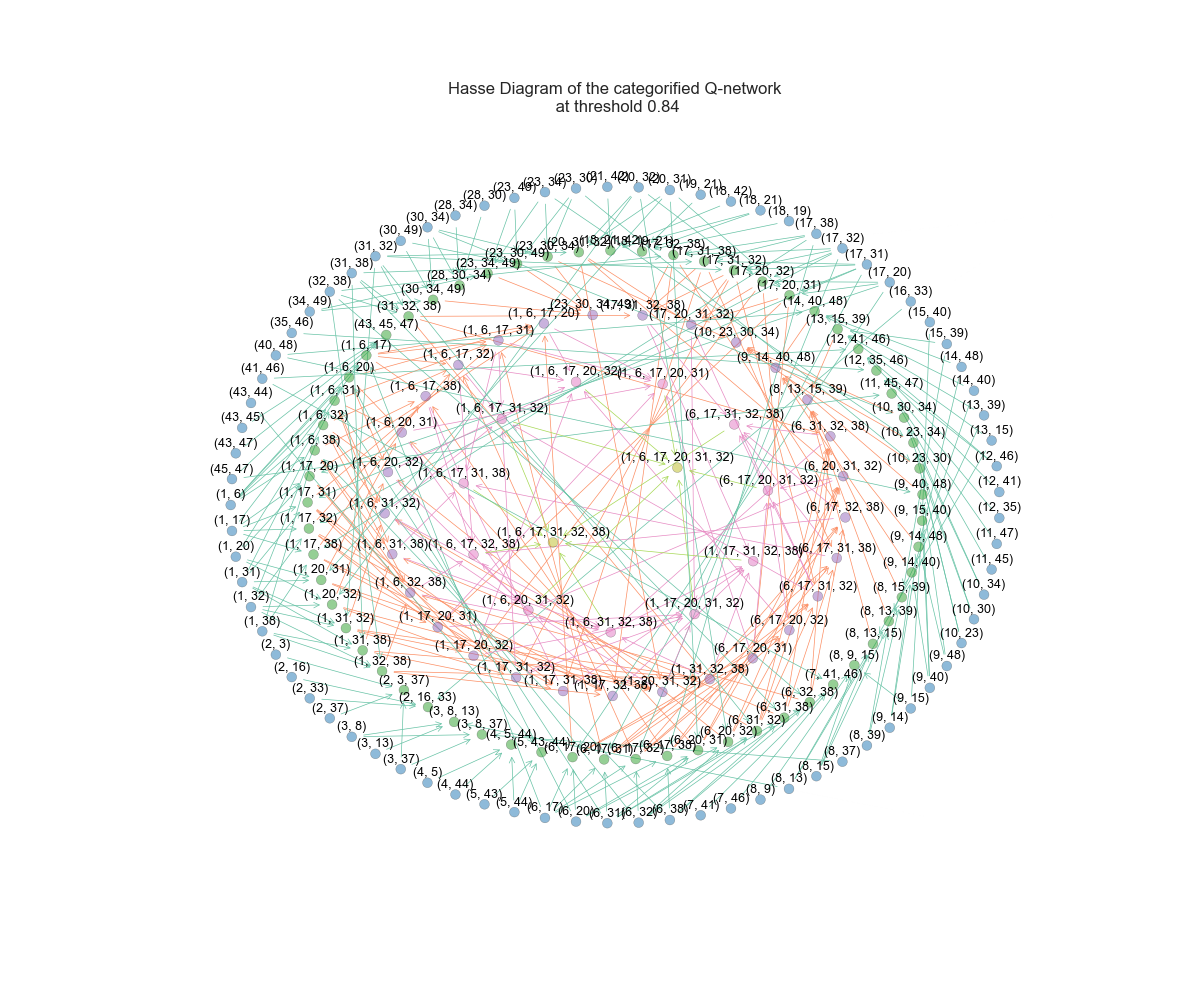} \\
        (a) Threshold $0.98$ & (b) Threshold $0.84$ \\
        \includegraphics[width=0.45\textwidth]{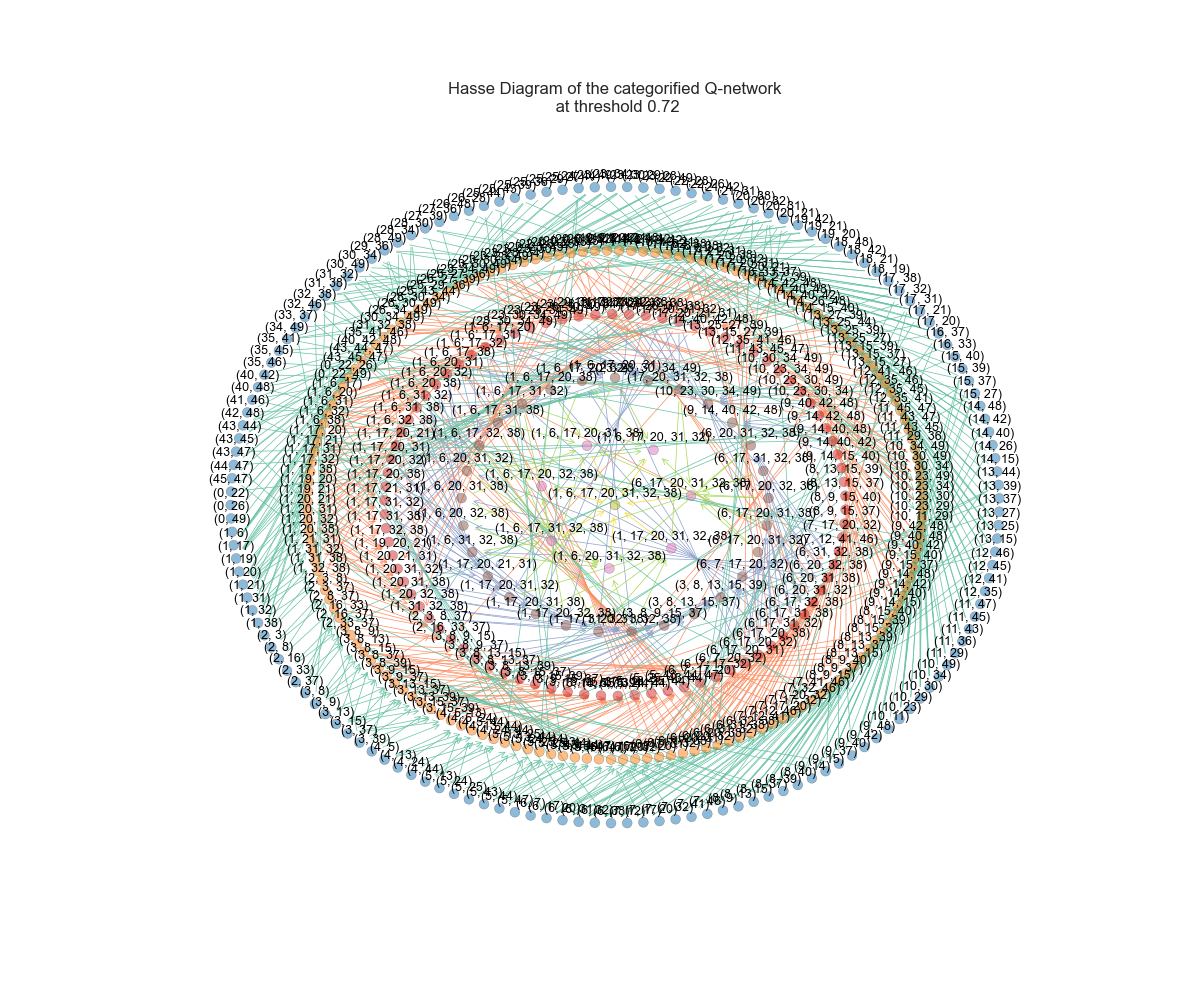} &
        \includegraphics[width=0.45\textwidth]{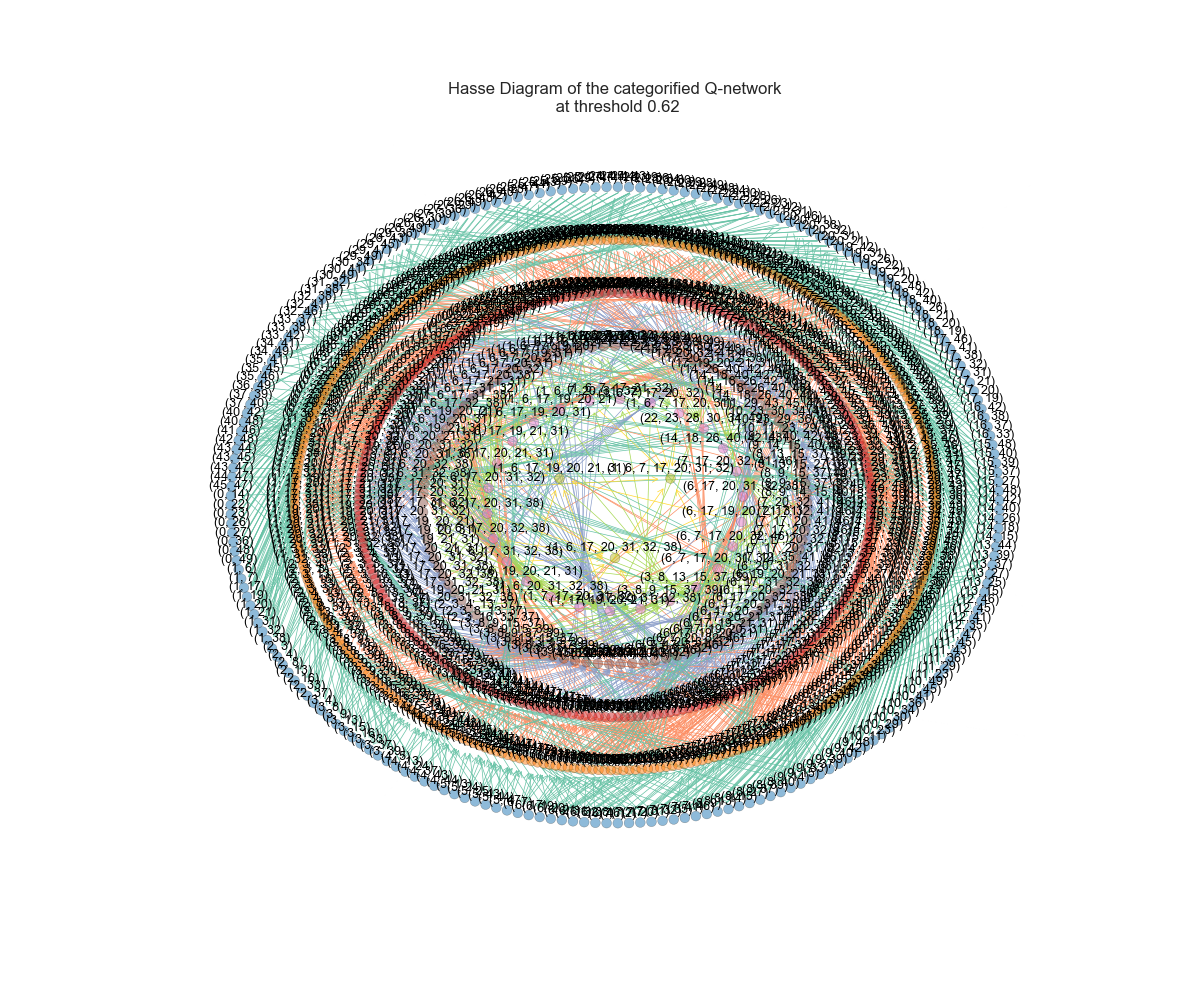} \\
        (c) Threshold $0.72$ & (d) Threshold $0.62$ \\
    \end{tabular}
    \caption{Hasse diagrams of \textcolor{magenta}{the inclusion-category defined over} the $\bm{Q}$-network at different thresholds. Each diagram shows minimal inclusions between cliques, treated as objects of the category. As the threshold decreases, the diagrams become increasingly dense and visually complex.}
    \label{fig:hasse}
\end{figure}

At this point, it should be emphasized that one cannot extract much quantitative information from visual inspection of these Hasse diagrams, especially at lower threshold values where the number of cliques and inclusions grows rapidly. The diagrams quickly become visually cluttered, obscuring structural features. For this reason, one \textcolor{magenta}{can} resort to computing topological invariants, such as Betti numbers $\beta_i$ derived from persistent homology, which provide concise and
interpretable summaries of the underlying invariants across thresholds. \textcolor{magenta}{Persistent homology is computed on filtrations of simplicial complexes. In this work we form the simplicial complexes via the clique (flag) construction: cliques in the underlying graph become simplices, and inclusions among cliques are recorded in an inclusion‑category whose nerve is the resulting simplicial complex. We do not compute persistent homology directly on arbitrary categorical constructions; this would, for example, run the risk of introducing spurious homological features because compositional morphisms create formal cycles. Appendix A summarizes the construction and its limitations.}

\begin{figure}[hbt]
    \centering
    \includegraphics[width=0.8\textwidth]{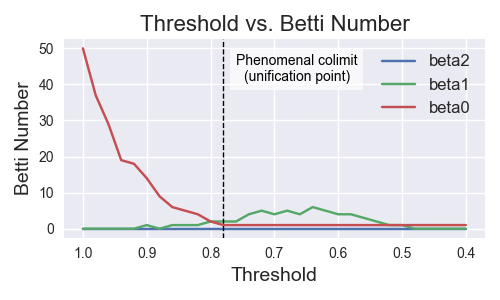}
    \caption{Betti numbers as a function of threshold. These topological invariants provide
    phenomenological interpretations of connectedness and cyclic structure. In the
    example, only $\beta_0$ (red) and $\beta_1$ (blue) take non-zero values, standing
    for the presence of disconnected components and loops, respectively. The dashed vertical line marks the colimit of the filtered category, i.e. the threshold at which all components unify into a single experiential field.}
    \label{fig:betti}
\end{figure}

Formally, $\beta_0$ counts connected components, while $\beta_1$ captures cycles. In our implementation, Betti numbers were computed for different thresholds (Fig. \ref{fig:betti}) using the \texttt{GUDHI} library \citep{gudhi14}, which provides efficient algorithms for building simplex trees and extracting persistent homology. 

\textcolor{magenta}{Phenomenologically,} we interpret $\beta_0$ as ``islands of experience,'' i.e. fragmented (micro-)contexts that remain disconnected at high thresholds. A high initial $\beta_0$ reflects a subjectivity scattered into many disjoint experiential units. As the threshold decreases, $\beta_0$ declines, indicating that these fragments begin to merge into larger experiential unities.

By contrast, $\beta_1$ \textcolor{magenta}{is interpreted as sign of} ``experiential loops,'' motifs where recurrent paths connect states. The rise of $\beta_1$ (peaking near threshold $\approx 0.65$ in our example) marks a phase of maximal relational tension: the experiential field is unified enough to sustain recurrent motifs, yet still diverse enough to generate ambiguity and multi-stability. When $\beta_0=1$ but $\beta_1>0$, the system exhibits a single experiential field enriched by internal cycles, suggesting a unified but internally complex subjectivity. As the threshold decreases further, $\beta_1$ falls again, reflecting the collapse of these tensions into higher-order
integration, where the experiential field becomes so interconnected that loops are absorbed into more global structures.

\textcolor{magenta}{As a minimal robustness check, one could verify that the qualitative behavior of $\beta_0$ and $\beta_1$ is stable under small perturbations of the similarity
threshold, invariant under relabeling, and consistent under alternative embeddings. Specifically, we checked that (i) the Betti curves are invariant under random permutations of the similarity matrix,  (ii) their qualitative behavior is robust when changing to a random linear projection, and (iii) the observed regime shifts in are not tied to a single fine‑tuned threshold but persist across short threshold intervals in the filtration.}

Higher Betti numbers \textcolor{magenta}{(not observed in the toy model)} admit further phenomenological speculation. A non-zero $\beta_2$ would correspond to ``experiential surfaces,'' enclosing relational flows in two-dimensional shells. A non-zero $\beta_3$ would indicate ``phenomenal bubbles,'' higher-order enclosures where tetrahedral relations wrap into sealed volumetric horizons. These higher-dimensional invariants would phenomenologically correspond to increasingly complex forms of experiential integration not reducible to simple connectedness or loops. It is an interesting question to what extent this points to a fundamental divergence in the phenomenal structure of, say, humans and future AI systems.

\subsubsection{Functorial actions, maximality, and colimits}

\begin{figure}[h]
    \centering
    \includegraphics[width=1\textwidth]{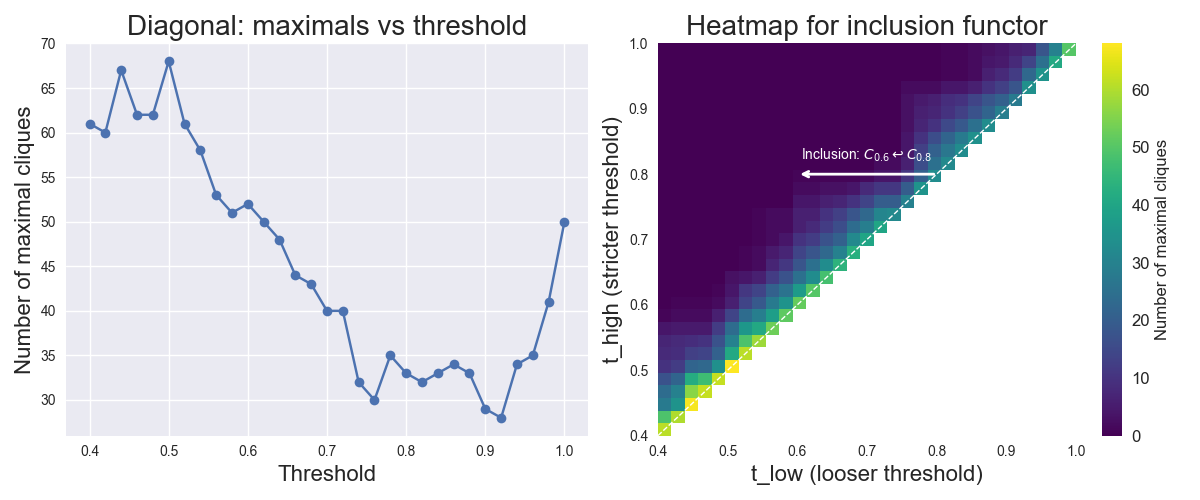}
    \caption{Diagonal curve (left) and heatmap (right) showing the number of maximal cliques under inclusion functors across thresholds.}
    \label{fig:inclusion}
\end{figure}

Finally, we \textcolor{magenta}{propose to represent} actions by functors on the category (implemented here as endomap on the cliques). Particularly simple examples are inclusion functor and \textcolor{magenta}{the functor obtained by restricting to the subcategory,}
corresponding to ``loosening'' or ``sharpening'' attention: 
\begin{equation}
\begin{aligned}
F_{\textrm{inc}} &: \mathcal{C}_{\textrm{high}} \hookrightarrow \mathcal{C}_{\textrm{low}}, 
&\qquad
F_{\textrm{res}} &: \mathcal{C}_{\textrm{low}} \rightarrow \mathcal{C}_{\textrm{high}}.
\end{aligned}
\end{equation}
Maximal cliques under these functors can be visualized in heatmaps (Fig. \ref{fig:inclusion}, only shown only for the inclusion functor).

While \textcolor{magenta}{these} functors describe how structures persist or disappear as attention is varied, we can also ask about the \textit{global} outcome of this process. Category theory captures this with the notion of a colimit: the minimal unifying object through which all clique‑inclusions (morphisms) factor. Phenomenologically, the colimit corresponds to the point where fragmented ``islands of experience” (high‑threshold cliques) converge into a unified experiential field.

Since we already computed Betti numbers, we can easily determine the threshold value where $\beta_0$ (connected components) collapses to 1, indicated by the dashed line in Fig. \ref{fig:betti}. This provides a concrete marker for the colimit, i.e. the moment when all inclusions factor through a single connected object. Additional phenomenological structure may be revealed by considering “local” colimits of subcategories, which appear as crossings of different Betti numbers $\beta_j$ and $\beta_k$ ($j\ne k$) cross.
 
\textcolor{magenta}{It is important to note that the examples discussed above
instantiate “action” only from the perspective of the analyst, via changes in the
similarity threshold that modulate the derived categorical structure. This should
not be confused with action in the enactive, agent–environment sense.\footnote{I
thank one reviewer for highlighting this distinction.} A genuinely agentic
implementation would require that the data itself be modified as a consequence of
the agent's actions, and that the categorical construction track these changes.}

\textcolor{magenta}{While developing such a setup lies beyond the scope of the present paper, we still like to speculate on a possible direction. In a biological or embodied artificial system, one
could record an “unperturbed” time series prior to action and a “perturbed” time
series after the agent intervenes in its environment. Provided that the
counterfactual case of “no action” does not itself induce comparable changes, the
difference between the corresponding categorical structures could serve as a
phenomenological marker of action. Another possibility is to train a predictive
model on the unperturbed time series (not unlike next‑state prediction in
current language models) and compare its predictions with the actual changes induced by
action. In both cases, the key idea is that actions would alter the relational
structure encoded by the $\bm{Q}$‑network, and the categorical machinery would then
capture how these alterations reorganize the agent's phenomenal interface.}

\subsubsection{Summary and Limitations}
The toy model demonstrates how third-person data can be systematically related to a first-personal structure, illustrating the phenomenological hierarchy (Fig. \ref{fig:hierarchy}): from raw features to phenomenal embeddings that define a $\bm{Q}$-network, and from networks to categories and invariants. While highly simplified, the model demonstrates how phenomenological properties can be mapped onto computational constructions, providing a bridge between the mental and the physical. \textcolor{magenta}{Still, the toy model is limited, especially insofar as it connects to the phenomenonological themes of self, time, and unity discussed in more details in the next Section \ref{sec:reflective}. This is not very surprising, after all the data used for this example was drawn from a synthetic time-series.}
\begin{itemize}
\item The collapse of $\beta_0$ was interpreted as \textcolor{magenta}{structural marker of phenomenological unification} in the toy model. By contrast, the notion of a \textit{self} \citep[whether ``pre-reflective'' or ``reflective'']{sep-self-consciousness-phenomenological}) requires more than mere connectedness: it presupposes a distinction between self and world, and potentially higher‑order reflective structures. While the present toy model does not implement these, its categorical framework suggests how such distinctions might be formalized in future extensions, \textcolor{magenta}{for example, by introducing a partition of the state space into ``self-directed'' and ``world-directed'' components and studying how functorial actions preserve or transform this partition. This would allow one to model the stability of self-related structures under changes of perspective or action.} 
\item \textcolor{magenta}{A further limitation concerns \textit{temporality}. In the toy model, temporal adjacency was removed before computing the categorical invariants. Retaining
temporal edges would have likely introduced artifacts.} Still, one could imagine reintroducing adjacency relations in a more realistic model as a way \textcolor{magenta}{to investigate a \textit{locus classicus} of phenomenology, namely} the temporal modes of consciousness \textcolor{magenta}{\citep{sep-consciousness-temporal}.}%
\item The \textit{unity} of consciousness can be related to constraints on how processes compose. In the toy model, we already speculated on the connection between attention and the inclusion functor. If consciousness is treated as a  processes, then unity corresponds to the requirement that the resulting diagrams cannot be rewritten into disconnected parts. This resonates with the idea of ``process‑unity'' that subsumes both synchronic and diachronic forms of unity under a single categorical condition.
\end{itemize}

\section{More complex phenomenological properties}
\label{sec:reflective}

\subsection{Self-consciousness}

In the previous sections, we argued that data about information-processing systems, both natural and artificial, can be integrated with the study of a first-person perspective via the notion of a $\bm{Q}$-network. Through \textcolor{magenta}{deriving categorical structures}, such $\bm{Q}$-networks can be studied with respect to (i) invariants, (ii) action-dependency, and (iii) \textcolor{magenta}{structural unification.}

\textcolor{magenta}{Beyond these relational features, there are further properties characteristic of the familiar, human form of} (“intentional”) consciousness. The minimal relational organization captured by a $\bm{Q}$‑network is therefore not yet intentional consciousness, but a structural precondition for it. Intentional consciousness arguably results from additional constraints that are imposed on a $\bm{Q}$‑network (see Fig.~\ref{fig:hierarchy}, top). 

\textcolor{magenta}{One important constraint pertains to the notion of ``self.''} On a conventional reading, the self is sometimes believed to be yet another type of representational content or a relation between those (e.g. the ``phenomenal model of the intentionality relation'' of \cite{Metzinger03b}). However, research into ``minimal phenomenal experience'' \citep{Metzinger20} or a literal reading of various “no-self” experiences, as they have been reported in multiple religious traditions, suggest that the self might not be necessary for consciousness \textit{as such}. In our model we understand self-consciousness, not as a necessary (substantial) entity, but as a structural feature of $\bm{Q}$-networks that allows an agent to distinguish its own states from environmental inputs. 
Phenomenologists often qualify \textcolor{magenta}{this further by assuming} the self to be a content of \textit{reflective} consciousness only. By contrast, the notion of a \textit{pre-reflective} self is not itself an object (or content) of consciousness but arises as structuring principle of intentional consciousness more generally. 

Here, category theory plays nicely alongside phenomenology and provides tools to translate qualitative statements into a mathematical language. 
For example, phenomenologists would typically  assume that the pre-reflective self could itself be reflected upon \citep{sep-self-consciousness-phenomenological}, i.e., made into the intentional object of a ``higher" act of consciousness. Reflection, in category theory, could be conceptualized in the form of a ``2-category'' \citep{nlab:2-category} – for example a “category of categories” – and the (reflective) self would be a (universal) object in this higher category.

To speculate even further, the more than 30 empirical notions of “self” \citep{Strawson97} could be given a more systematic reading as correlates of an ``$n$-self", which refers to a universal construction on an $n$-category. Moreover, there is a natural sense in which $n$-selves can be nested, though this need not be the case, cf. the intricacies of ``ramification" \citep{Ehresmann07}.

\subsection{Time-consciousness}

The relations giving rise to self-consciousness are orchestrated throughout the \textcolor{magenta}{dynamical} evolution of a system. Metaphorically speaking, time is the glue between the self and the world. Perhaps, time-consciousness has been the most studied concept from phenomenology to date, which has attracted various empirical and computational studies, spanning neuroscience, predictive processing, and even psychopathology research \citep{Wiese17b,Kent21,Singhal21,Tewes20}.

Phenomenologists have long speculated on the way how time-consciousness could be described at least since \citep{Husserl66}. Importantly, describing time-consciousness is not tantamount to describing the subjective experience of time in terms of, for example, ``past'', ``present'' and ``future'' (i.e., tensed time). These are secondary aspects, ``experienced labels,'' that we attach to particular instances of consciousness (and their contents). In contrast, time-consciousness refers to an organizing principles of the broader dynamics of intentional consciousness itself. 
It is natural to interpret Husserl's tri-partite model of ``retention,'' ``primal impression,'' and ``protention'' \citep{sep-consciousness-temporal} in terms of a $\bm{Q}$-network that supports three systematically interrelated families of indexed functors. \textcolor{magenta}{For example, in section \ref{subsec:toy-models}, we separated temporal adjacency from similarity-based structures. One could now compare invariants computed on (i) a similarity-only graph and (ii) a temporally constrained graph (for example, restricting edges to a fixed lag window), thereby distinguishing structural features that are topological from those that are temporal. Such a comparison would be essential in applications to biological or embodied systems beyond simple toy models.}

\textcolor{magenta}{More generally}, an investigation into time-consciousness would likely stay non-reductive. While we seem to know how time works in (classical\footnote{There exist serious puzzles in understanding time in fundamental physics \citep{Price96,Brukner14}.}) physics, e.g., how chemical reactions occur over a measurable period of time or how neurons fire relative to an external clock, one is hard-pressed to account for a basic puzzle in the philosophy of time: how are genuine change and its experience at all possible? Relatedly, phenomenologists typically try to stop presupposing an external, physical world. Even in this (perhaps only imaginary) situation, phenomenologists argue that one's experience exhibits a sense of “time,” which is remarkably distinct from the observable  physical dynamics. This issue has been probed repeatedly in neurophenomenology, starting with the seminal works featured in \citep{Petitot99}. The state of the art approach often involves a dual description of physical (neuronal) and psychological systems, informed by phenomenological principles, see also \citep{Horvat22,Singhal24} for a recent take on this. \textcolor{magenta}{Distinguishing between the temporal profile of experience as it is given by physical transitions of the systems (mirrored by temporal adjacencies in the $\bm{Q}$-network) and the phenomenological aspects that are given by other structural features (e.g. mirrored in the similarity-structure of the graph) would be consistent with this dual approach.}

\subsection{Process Unity}

What further distinguishes \textcolor{magenta} {the minimal relational organization of} $\bm{Q}$-networks from intentional forms of (e.g., human) consciousness is encoded by constraints on the latter such as the unity of consciousness \citep{Bayne10}. \textcolor{magenta}{Spelling out this unity-constraint amounts to a major formal research question in the science of consciousness \citep{Bayne03,Tononi16,Prentner19a,Kleiner24b}.}

To \textcolor{magenta}{translate this into the conceptual language of the current work} we appeal to applied category theory \citep{Abramsky08,Bradley18}. In applied category theory, objects are often called ``types'', which are taken to represent, in our case, \textcolor{magenta}{categorical structures derived from} $\bm{Q}$-networks. In addition, applied category theory knows of ``processes'' that operate upon these types. Each process would then refer to a transformation of \textcolor{magenta}{these structures.}

\begin{figure}[htb]
\centering
\begin{tabular}{ccc}
\begin{minipage}{0.3\textwidth}
$
\xymatrix{
\ar[d]_t &&\ar@{-}[d]_{X} &\\
&&*++[F]{f} \ar@{-}[d]_{Y}\\
&&*++[F]{g} \ar@{-}[d]_{Z} \\
&&&}
$
\end{minipage} 
& \begin{minipage}{0.125\textwidth} $\Leftrightarrow$ \end{minipage}& 
\begin{minipage}{0.225\textwidth}
$
\xymatrix{\ar@{-}[d]_<{X} &&\\
 *++[F]{f} \ar@{-} `d[rd]`[r]_Y[r]  & *+[F]{g^T} \ar@{-} `u[ru]`[r] [dr]^>{Z} &  \\
&&& }
$
\end{minipage}
\end{tabular}
\caption[String diagram]{Generic string diagram for a strict monoidal category illustrating the idea of ``process-unity." Types (wires; experiences described by categorical structures derived from $\bm{Q}$-networks) are transformed by processes (boxes; interpreted as transformations of these structures). On the left side, one sees a sequential composition of processes ($g \circ f: X \rightarrow Z$). On the right, is an equivalent diagram with the respective processes not composing sequentially but horizontally ($  m \circ (f \otimes g^T \otimes 1) \circ p: X \otimes 1 \rightarrow 1 \otimes Z$)\footnotemark; time flows from top to bottom.}
\label{fig:string}
\end{figure}
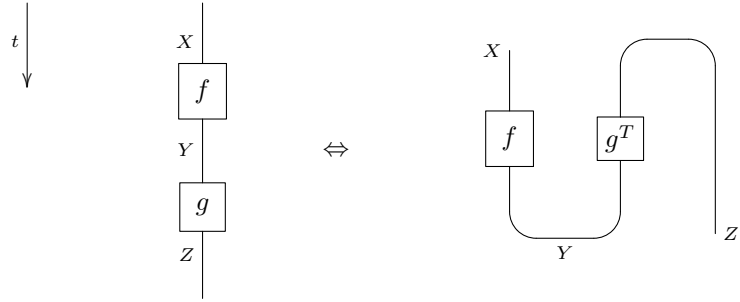

Applied category theory studies the \textit{composition of processes} \citep{Coecke21}, which can be naturally interpreted as a single process. 
Graphically, this is represented with a string diagram, 
where \textcolor{magenta}{categorical structures derived from} $\bm{Q}$-networks (types) are transformed by processes (boxes), and sequential or parallel compositions yield new processes (Fig.~\ref{fig:string}).  

%
\footnotetext{\textcolor{magenta}{Here “$\circ$” denotes sequential composition of morphisms, while “$\otimes$” denotes parallel composition.} $T$ denotes the transpose; $m$ and $p$ are horizontal composites of the identity and other generic objects (``cups'' and ``caps'' \citep{Signorelli20a}).
}
In applied category theory, emphasis is often placed on the equivalence of string diagrams according to so-called ``re-writing rules.'' If one diagram can be re-written in terms of a second, i.e., if there exist inverse (\textcolor{magenta}{up to natural transformation}) functors between them, they are equivalent. This lets us represent the idea of constraints as \textcolor{magenta}{categorical} equivalences on compositions of processes \citep{nlab:equivalence_of_categories}. It is here where we see the main relation to the unity of consciousness.
In consciousness studies, one often distinguishes a “diachronic” unity of consciousness across time and a “synchronic” unity at the same point in time \citep{Dainton00}. But believing that either a diachronic or synchronic sense of unity is in some way more fundamental might stem from a conflation of single \textcolor{magenta}{states of} experiences (which are trivially unified) with intentional forms of consciousness (which can potentially be dis-unified). Following \citep{Signorelli21b}, it might be better to subsume both under the broader notion of a “process unity”: intentional consciousness is unified iff its representation as a string diagram cannot be written in an equivalent but disconnected way; compare Fig. \ref{fig:string}.

\section{Conclusions and outlook}
\label{sec:conclusion}

In this article, we proposed a \textcolor{magenta}{phenomenological framework for computational systems.} 
We first argued on \textcolor{magenta}{conceptual grounds} that category theory is suitable for understanding phenomenology's “doubly relational” nature ($\Phi = R^2$). We then outlined an exemplary way to base such a treatment on $\bm{Q}$-networks, specifically claiming that some central notions of phenomenology can be represented \textcolor{magenta}{by categorical objects derived from $\bm{Q}$-networks.}  

Category theory appears promising not only for its relational character but also for the precision of its methods. We hypothesize that the information-processing in various systems \textcolor{magenta}{(including AIs) can} be embedded in $\bm{Q}$-networks \textcolor{magenta}{that define} a first-person perspective insofar as they are organized around concepts that articulate phenomenological themes such as intentionality, subjectivity, etc. \textcolor{magenta}{Category theory serves as a high-level link that facilitates interpretation and design of those networks.}

\textcolor{magenta}{What could falsify this? For example, one could examine whether features such as ``structural unification'' (Section \ref{subsec:deriving}) are present or absent in systems taken to be conscious. But this immediately raises a problem. Either we rely on intuitive judgments about which systems are conscious, which is highly problematic, since consciousness is not a third-personally observable phenomenon.\footnote{\textcolor{magenta}{This is reminiscent of the unresolved debate about ``conscious grids'' in IIT \citep{Merker22,Grasso21}. Some might find this so counter-intuitive so as to count as as falsification of the theory. Others find this to be a major predictive strength of the formalism.}} Alternatively, we compare observable behavior with scientific determinations of consciousness \citep{Bayne24}.} Yet, especially when it comes to conscious AI and the question of moral status \citep{Shevlin24}, this is problematic too! For some systems, such as vertebrates, this can perhaps be done using a “theory-light” approach \citep{Birch22}. However, machines could conceivably produce sophisticated behavior without consciousness being involved. 

One might thus think that without a good theory that explains the (metaphysical) relation between physical states and consciousness, we will ultimately not know whether conscious AI has a “realistic possibility of conscious experience”  \citep{declaration}. \textcolor{magenta}{However,} this presents the ultimate problem for empiricist approaches to consciousness: nothing in our experience seems to show that something else is \textcolor{magenta}{or is not} a locus of experience. At the same time, this suggests \textcolor{magenta}{the need for} a novel reasoning strategy beyond received empiricist approaches, no matter whether they are “theory-light” or “theory-heavy” \citep{Butlin23} or \textcolor{magenta}{whether they  define ``theory-derived'' indicators of consciousness \citep{Butlin25}}. 

\begin{figure}[htb]
\begin{equation}
\nonumber
\xymatrixrowsep{.5cm}
\xymatrix{
& *+[F-:<3pt>]{\textrm{phenomenological concept}} \ar@{.>}@/^/[ddl]\ar@{.>}@/^/[dd]\ar@{.>}@/^/[ddr]^{\textrm{constrains}}  \\
& & & \\
*+[][F-]{x_0} \ar@{-}[u] \ar@{-}[d] &  *+[][F-]{x_1} \ar@{-}[u] \ar@{-}[d]& *+[][F-]{x_2} \ar@{-}[u] \ar@{-}[d]\\
& & & 
} 
\end{equation}
\caption{Schematic illustration of how individual computational processes are organized under a phenomenological concept. The concept functions as a constraint on processing, not as another layer of computation, as indicated by the dotted arrows. \textcolor{magenta}{This jointly defines the phenomenological structure of the interface instantiated by a computational system.}}
\label{fig:transcendental}
\end{figure}
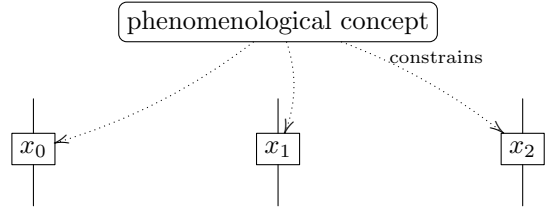

\textcolor{magenta}{Instead, we recommend a different strategy.} In his \textit{Prolegomena}, Immanuel \cite{Kant04} asked how mathematical knowledge is possible, for example, how we can know that the angles of a triangle sum to 
$180^\circ$. 
\textcolor{magenta}{He argued that we can know such things because our minds are set up to experience the world in terms of space. For Kant, space is the basic format built into how we sense anything.} 
By analogy, we may ask: how is subjective experience possible?
Our proposal is that computational processes, when subsumed under phenomenological concepts, are thereby constrained in ways that make subjective experience possible 
(Fig. \ref{fig:transcendental}). \textcolor{magenta}{These constraints jointly determine the \textit{format of the interface} through which a system enacts its first-person perspective, and thus relates to the world.} 

We therefore suggest focusing on \textcolor{magenta}{evaluating} the relational richness of the interfaces that could be instantiated by various systems, such as biological ones or AIs. \textcolor{magenta}{The structural markers proposed here, such as the degree of structural unification (in the toy model revealed by topological invariants) or the way categorical structures reorganize under action‑indexed transformations (self- and time-consciousness or process-unity), offer concrete means of assessing how richly an interface relates its possible states to one another. They do not operationalize consciousness itself, but they reveal whether a system’s relational architecture exhibits the kinds of multi‑level coherence that phenomenology requires.}\footnote{\textcolor{magenta}{In a different context, this has recently been operationalized as ``SLP-tests'' that test for the subjective-linguistic, latent-emergent, and phenomenological-structural dimensions of AI-interfaces \citep{Prentner25ac}.}} 

Future work should extend this analysis of $\bm{Q}$-networks to different kinds of third‑person data — molecular, cellular, neuronal, or digital — in order to test how far the categorical framework generalizes across levels of organization. Such comparative analyses would help clarify whether the same phenomenological concepts can constrain processes at multiple scales, and whether artificial systems can be meaningfully embedded into this scheme.

Consciousness is not an event in the world; it is the event in which a world appears.


\newpage
\section*{Appendix A: Persistent Homology, Clique Complexes, and the Inclusion-Category}
\label{appA}
The aim of this appendix is to make explicit how to compute persistent homology, which is standardly defined for filtrations of simplicial complexes, and show how this is related to the inclusion-category used to connect this to phenomenology. 
 
The standard way to derive a simplicial filtration from a $\bm{Q}$-network goes as follows:
\begin{enumerate}
\item The \textbf{underlying graph $G$} is given by the $\bm{Q}$-network, which resulted from embedding physiological data (in the toy model: a synthetic timeseries) into a phenomenal space; vertices correspond to embedding points, edges correspond to cosine-similarities between embeddings in the phenomenal space.
\item \textbf{Enumerate all cliques $C$} (up to a chosen maximum size if needed). A clique is defined as a set of vertices that are pairwise connected by edges. The clique complex (also called the flag complex) of a graph replaces every $k$-clique by a $(k-1)$-simplex \citep{Edelsbrunner10}.
\item A nested sequence of simplicial complexes $K_0 \subset K_1 \subset \hdots \subset K_T$ is obtained by varying a scale or threshold parameter. In our case, the threshold parameter is defined with respect to the cosine-similarities. The \textbf{filtration} $\lbrace K_t \rbrace$ is obtained by lowering the similarity threshold, thereby adding edges and hence new cliques.
\item \textbf{Persistent homology} is computed by evaluating homology groups $H_k(K_t)$ across the filtration and records the birth and death of homological features (connected components, loops, voids) as intervals (persistence). The output is typically barcodes or persistence diagrams. For our toy-model, we recorded the $i$-th Betti-number to map persistent features back to domain semantics (e.g., long‑lived 1‑cycles indicate robust loop structures among cliques.)
\end{enumerate}

The inclusion‑category is a categorical rephrasing of the simplicial complex structure.\footnote{Rather than, for example, a free categorical construction that would possibly introduce formal compositional cycles unrelated to geometric simplices.} It encodes the same combinatorial data as the clique complex: every object corresponds to a simplex, and inclusions correspond to face relations. Because morphisms are inclusions, there is a canonical nerve/simplicial realization: the nerve of the inclusion‑category is (isomorphic to) the clique complex. This gives a direct, canonical route to persistent homology. 

While long persistence suggests robustness, not every persistent feature implies a direct phenomenological or causal structure without further validation. Moreover, if future work introduces richer categorical morphisms (beyond mere inclusions), then the nerve construction may no longer capture the intended semantics and a careful theory of persistent homology on such categories would be required. That is a substantial mathematical project that goes outside the present scope.

\backmatter

\end{document}